\documentclass[5p]{elsarticle}

\usepackage{graphicx}
\graphicspath{{figure/}}
\DeclareGraphicsExtensions{.eps, .pdf}
\usepackage[caption=false,font=footnotesize,labelfont=sf,textfont=sf]{subfig}
\usepackage{picinpar}
\usepackage{color}		
\usepackage{amsmath,amssymb,amsfonts}
\usepackage{url}		
\usepackage{bm}			
\usepackage{booktabs}
\usepackage{multirow}
\usepackage{balance}
\usepackage{newtxtext}
\usepackage{newtxmath}
\usepackage{algorithm}
\usepackage{algorithmic}
\usepackage{lineno}
\usepackage[colorlinks, bookmarks, unicode]{hyperref}

\modulolinenumbers[5]
\journal{}

\newcommand{\trace}[1]{\text{tr}({#1})}
\newcommand{\norm}[1]{||{#1}||}
\newcommand{\transpose}{{\text{T}}}
\newcommand{\dimension}[1]{\in\mathbb{R}^{#1}}
\newcommand{\fro}{{\text{F}}}

\newtheorem{theorem}{Theorem}

\begin{document}

\begin{frontmatter}

\title{Double Weighted Truncated Nuclear Norm Regularization \\ for Low-Rank Matrix Completion}

\author{Shengke~Xue\corref{corr1}}
\cortext[corr1]{Corresponding author}
\ead{xueshengke@zju.edu.cn}

\author{Wenyuan~Qiu\corref{}}
\ead{qiuwenyuan@zju.edu.cn}

\author{Fan~Liu}
\ead{flyingliufan@zju.edu.cn}

\author{Xinyu~Jin}
\ead{jinxinyuzju@gmail.com}

\address{College of Information Science and Electronic Engineering, Zhejiang University, \\ No.~38 Zheda Road, Hangzhou 310027, China}

\begin{abstract}
Matrix completion focuses on recovering a matrix from a small subset of its observed elements, and has already gained cumulative attention in computer vision. Many previous approaches formulate this issue as a low-rank matrix approximation problem. Recently, a truncated nuclear norm has been presented as a surrogate of traditional nuclear norm, for better estimation to the rank of a matrix. The truncated nuclear norm regularization (TNNR) method is applicable in real-world scenarios. However, it is sensitive to the selection of the number of truncated singular values and requires numerous iterations to converge. Hereby, this paper proposes a revised approach called the double weighted truncated nuclear norm regularization (DW-TNNR), which assigns different weights to the rows and columns of a matrix separately, to accelerate the convergence with acceptable performance. The DW-TNNR is more robust to the number of truncated singular values than the TNNR. Instead of the iterative updating scheme in the second step of TNNR, this paper devises an efficient strategy that uses a gradient descent manner in a concise form, with a theoretical guarantee in optimization. Sufficient experiments conducted on real visual data prove that DW-TNNR has promising performance and holds the superiority in both speed and accuracy for matrix completion.
\end{abstract}


\end{frontmatter}


\section{Introduction} \label{sec:introduction}
%
%
%
%

Matrix completion tries to estimate missing elements of an incomplete matrix with only part of them observed, which remains a valuable challenge in computer vision. This problem originates from online recommendation systems, and has become increasingly attractive in various researches, e.g., motion capture \cite{Hu2017-Motion, Adeli2015-Non-negative}, image recovery \cite{Li2015-Non-Local, Hu2013-Accurate}, image classification \cite{Luo2015-Multiview}, background subtraction \cite{Mansour2014-Video, Yang2014-Background}, and dynamic imaging \cite{Lee2016-Computationally}.

Matrix completion \cite{Candes2010-Matrix, Candes2010-Power} preserves a low-rank or approximately low-rank structure of the restored matrix when estimating the missing values from a partial sampling of the observed data. Mathematically, given an incomplete matrix $\mathbf{M} \dimension{m \times n}$, it usually can be formulated as
\begin{equation}
\min_{\mathbf{X}} \ \text{rank}(\mathbf{X}) \quad \text{s.t.} \ \mathbf{X}_{ij} = \mathbf{M}_{ij},\ (i,j) \in \Omega \, ,
\end{equation}
where $\Omega$ is a set of locations in matrix $\mathbf{X}$, corresponding to the known entries. 

Unfortunately, the rank$(\cdot)$, non-convex and discontinuous in nature, is NP-hard generally. The rank minimization cannot be directly solved with efficiency. However, Cand\`{e}s and Recht \cite{Candes2009-Exact} initially proved that the nuclear norm minimization is able to exactly recover a matrix from adequate observed elements, provided that the incoherence condition is satisfied \cite{Liu2016-Low-Rank}. Thus, the nuclear norm is widely adopted as a convex surrogate to the rank function.

However, existing nuclear norm based approaches, such as the robust principal component analysis (RPCA) \cite{Candes2011-Robust, Wright2009-Robust} and the singular value thresholding (SVT) method \cite{Cai2010-Singular}, obtain sub-optimal solutions in reality and entail a number of iterations to converge. Since the nuclear norm may not be an appropriate substitute for the rank function and the theoretical requirements of the nuclear norm heuristic are usually violated in practice. Thus, the truncated nuclear norm regularization (TNNR) approach \cite{Hu2013-Accurate} is presented as an accurate and robust approximation for the rank function. A two-step alternating scheme is adopted for the updating procedures. It designs an alternating direction method of multipliers (ADMM) \cite{Lin2011-Linearized} and an accelerated proximal gradient line search (APGL) method \cite{Toh2010-accelerated} to iteratively solve a convex sub-problem in the second step of the TNNR. 

Though the accuracy of reconstruction has been obviously improved by the TNNR method, the speed of convergence is not noticeably promoted. Besides, the TNNR method is not robust to the parameter $r$ (the number of truncated singular values). It becomes worse in certain cases with an inappropriate $r$, which induces that the TNNR method fails in some real applications.

However, TNNR attempts to recover all missing entries of an incomplete matrix simultaneously in each step. Intuitively, the task of restoring only a few number of lost elements is easy. In other word, the matrix completion problem will be plausible when the absent components are recovered orderly, i.e., from easy parts to difficult parts. To this end, a weighting manner can be integrated in optimization. This paper proposes a double weighted truncated nuclear norm regularization (DW-TNNR) method. Different from the TNNR that deals with all rows and columns of the target matrix equally, our proposed approach assigns different weights to some rows and columns separately, based on the number of known elements in the corresponding dimensions. It establishes a priority to the missing values of an incomplete data. By recovering these elements sequentially, the easy parts (with a small weight) will be roughly recovered first. Subsequently, the entire matrix can be restored efficiently and the whole task can be apparently accelerated. 
In addition, \cite{Lu2016-Nonconvex, Li2017-Weighted} both verified that the weighting scheme can be integrated in low-rank representation.

Although the ADMM and APGL both converge with theoretical guarantees, they have to go through a vast number of iterations, which results in considerable time consumption. To further accelerate the convergence speed of the TNNR method, we derive a closed form solution by the gradient descent method and avoid solving an objective function iteratively in the second step of TNNR. Beneficial from a gradient descent scheme, our proposed method is able to effectively solve the corresponding sub-problem in one-step fashion. In addition, the gradient descent method was recently involved in low-rank representation \cite{Song2016-Image, Xue2017-Robust}. Though converging to a local minimum finally, the gradient descent method is more efficient than the ADMM and APGL, and it obtains acceptable results empirically.

The major contributions of this paper include:
\begin{itemize}
	\item DW-TNNR deploys different weights to a deficient data for the acceleration of convergence. It conforms the rule that some rows and columns with more known elements, respectively, are restored with higher priority and accuracy than the others. Specifically, some lost elements will be recovered prior than the others, if associated with a smaller weight. 	
	
	\item This paper designs an efficient gradient descent method theoretically, with the assurance of local convergence, in a concise formulation rather than the iterative optimization fashion in the second step of TNNR. Experiments confirm that our DW-TNNR runs significantly faster than the compared approaches.
	
	\item DW-TNNR is more robust to the number of truncated singular values than some existing approaches. Evidence indicates that it is applicable to various scenarios of the element loss in matrix completion.	
\end{itemize}



\section{Related Work} \label{sec:related_work}

Matrix completion \cite{Bhojanapalli2014-Universal, Chen2014-Coherent} gradually attracts considerable interests in various areas. For videos with mobile background, Mansour and Vetro \cite{Mansour2014-Video} compensated it for the variation in the camera perspective, by using the motion vectors extracted from the coded video bitstream. Likewise, Yang \emph{et~al.}~\cite{Yang2014-Background} proposed a motion-assisted matrix completion to allocate the reliability of pixels from background. The iteratively reweighted nuclear norm (IRNN) method \cite{Lu2016-Nonconvex} solved a weighted singular value thresholding problem, by taking the weight vectors as the gradient of concave regularizations. Though non-convex and non-smooth, the IRNN has a closed form solution.

Instead of standard nuclear norm, some researches devise the variants of nuclear norm to improve its performance. Oh \emph{et~al.}~\cite{Oh2016-Partial} proposed the partial sum minimization of singular values (PSSV) method, to replace the traditional nuclear norm in RPCA. It implicitly expected a soft constraint of the target rank. The objective of the PSSV is given by
\begin{equation}
\min_{\mathbf{A},\mathbf{E}} \ \norm{\mathbf{A}}_{p = N} + \lambda \norm{\mathbf{E}}_1, \quad \text{s.t.} \ \mathbf{O} = \mathbf{A} + \mathbf{E},
\end{equation} 
where $N$ is the target rank of $\mathbf{A}$, and $\norm{\cdot}_1$ is the $\ell_1$ norm. However, $N$ entails to be determined before optimization and $N$ is different based on various scenarios. Moreover, this objective is highly non-convex.

Similarly, the joint Schatten-$p$ norm and $\ell_p$ norm \cite{Nie2015-Joint} was used to substitute the rank function and enhance the robustness to outliers. The $\ell_p$ norm of a vector $\mathbf{v}$ is defined as $\norm{\mathbf{v}}_p = (\sum_i |\mathbf{v}_i|^p)^{\frac{1}{p}}$. The definition of  Schatten-$p$ norm ($0 < p < \infty$) of a matrix $\mathbf{X} \dimension{m \times n}$ is
\begin{equation}
	\norm{\mathbf{X}}_{S_p} = \left ( \sum_{i=1}^{\min(m,n)} \sigma_i^p \right )^{\frac{1}{p}} ,
\end{equation}
where $\sigma_i$ is the $i$-th singular value of $\mathbf{X}$. Although the Schatten-$p$ norm can approximate the nuclear norm ($p=1$) and the rank ($p = 0$) respectively, the objective function of which is non-convex. It takes the alternating direction method for optimization, which is not efficient enough.

The TNNR was first presented by Hu \emph{et~al.}~\cite{Hu2013-Accurate}, which approximated to the rank function better than the nuclear norm. In contrast to treating all singular values together, the TNNR leaves out the largest $r$ singular values, and tries to minimize the smallest $\min(m,n)-r$ singular values, where $m$, $n$ are the dimensions of data, and $r$ is the number of truncated singular values.
In advance, let us define 
\begin{equation}
(\mathbf{M}_{\Omega})_{ij} = \begin{cases}
\mathbf{M}_{ij}, & (i,j) \in \Omega \, , \\
0, & (i,j) \in \Omega^\text{c},
\end{cases}
\end{equation}
where $\Omega$ is the set of locations in $\mathbf{M}$ with respect to the known elements and $\Omega^\text{c}$ is the counterpart with respect to the missing elements. It indicates that $\Omega^\text{c}$ is the complement of $\Omega$. Hence, the TNNR minimization model is formulated as
\begin{equation}  \label{eq:TNNR_objective}
\begin{aligned}	
\min_{\mathbf{X}} \ & \norm{\mathbf{X}}_* - \max_{\mathbf{C}\mathbf{C}^\transpose=\mathbf{D}\mathbf{D}^\transpose=\mathbf{I}} \trace{\mathbf{C}\mathbf{X}\mathbf{D}^\transpose} \\
\text{s.t.} \ & \ \mathbf{X}_{\Omega} = \mathbf{M}_{\Omega},
\end{aligned}
\end{equation}
where $\mathbf{C} \dimension{r \times m}$, $\mathbf{D} \dimension{r \times n}$, $\norm{\,\cdot\,}_*$ denotes the nuclear norm (sum of all singular values), and $\trace{\cdot}$ denotes the trace function. It can be solved by a two-step iterative fashion. Let $\mathbf{X}_1=\mathbf{M}_\Omega$ as initialization. In Step 1 of the $\ell$-th iteration, assume $\mathbf{U}_\ell \Sigma_\ell \mathbf{V}_\ell^\transpose$ is the singular value decomposition (SVD) of $\mathbf{X}_\ell$, where $\mathbf{U}_\ell \dimension{m \times m}$ and $\mathbf{V}_\ell \dimension{n \times n}$ are the left and right orthogonal matrices. Thus, $\mathbf{C}_\ell$ and $\mathbf{D}_\ell$ are calculated directly through 
\begin{equation}
\mathbf{C}_\ell = [\mathbf{u}_1, \cdots, \mathbf{u}_r]^\transpose,\ \mathbf{D}_\ell = [\mathbf{v}_1, \cdots, \mathbf{v}_r]^\transpose.
\end{equation}

In Step 2, $\mathbf{X}_{\ell+1}$ is obtained by solving the following sub-problem:
\begin{equation} \label{eq:TNNR_step_2}
\begin{aligned}	
\min_{\mathbf{X}} \ & \norm{\mathbf{X}}_* - \trace{\mathbf{C}_\ell\mathbf{X}\mathbf{D}_\ell^\transpose} \\
\text{s.t.} \ & \ \mathbf{X}_{\Omega} = \mathbf{M}_{\Omega}.
\end{aligned}
\end{equation}
Two typical optimization approaches were developed in \cite{Hu2013-Accurate} to minimize \eqref{eq:TNNR_step_2}, i.e., the alternating direction method of multipliers and the accelerated proximal gradient line search method. However, both of them require numerous iterations to converge and obtain sub-optimal results in Step 2. The TNNR algorithm alternately executes the two steps above. 

Recently, a variety of studies were derived from the TNNR. Hong \emph{et~al.}~\cite{Hong2016-Online} combined the truncated nuclear norm with the online RPCA \cite{Feng2013-Online}, to promote low dimensional subspace estimation. Motion capture data completion \cite{Hu2017-Motion} demonstrated the validity via integrating it with the truncated nuclear norm. Large scale multi-class classification which uses the TNNR and multinomial logistical loss was suggested in \cite{Hu2015-Large}. Lee and Lam \cite{Lee2016-Computationally} applied the truncated nuclear norm heuristic to ghost-free high dynamic range imaging by searching the low-rank structure of irradiance maps. Cao \emph{et~al.}~\cite{Cao2017-Recovering} extended the TNNR to the low-rank and sparse decomposition problem, and applied it for foreground object detection.

\section{Double Weighted Truncated Nuclear Norm Regularization} \label{sec:proposed_method}
\subsection{Problem Formulation}


On the basis of the Von~Neumann's trace inequality \cite{Neumann1937-matrix}, for any given matrices $\mathbf{X} \dimension{m \times n}$, $\mathbf{A} \dimension{m \times m}$, and $\mathbf{B} \dimension{m \times n}$, we have the property that
\begin{equation} \label{eq:von_neumann_trace_inequality}
\trace{\mathbf{A}\mathbf{X}\mathbf{B}^\transpose} \leq \norm{\mathbf{X}}_* ,
\end{equation}
where $\mathbf{A}\mathbf{A}^\transpose=\mathbf{I}$ and $\mathbf{B}\mathbf{B}^\transpose=\mathbf{I}$. Hence, the minimization formulation \eqref{eq:TNNR_objective} is rewritten as follows: 
\begin{equation} \label{eq:rewritten_TNNR_objective}
\begin{aligned}
\min_{\mathbf{X}} \ & \max_{\mathbf{A}\mathbf{A}^\transpose=\mathbf{B}\mathbf{B}^\transpose=\mathbf{I}} \trace{\mathbf{A}\mathbf{X}\mathbf{B}^\transpose} - \max_{\mathbf{C}\mathbf{C}^\transpose=\mathbf{D}\mathbf{D}^\transpose=\mathbf{I}} \trace{\mathbf{C}\mathbf{X}\mathbf{D}^\transpose} \\
\text{s.t.} \,\ & \quad \mathbf{X}_{\Omega} = \mathbf{M}_{\Omega}.
\end{aligned}
\end{equation}
Here, $\mathbf{A} \dimension{m \times m}$, $\mathbf{B} \dimension{m \times n}$, $\mathbf{C} \dimension{r \times m}$, and $\mathbf{D} \dimension{r \times n}$, all of which are orthogonal matrices.

According to \eqref{eq:rewritten_TNNR_objective}, we first design a two-step iterative scheme. Assign $\mathbf{X}_1 = \mathbf{M}_{\Omega}$ as initialization. In the $k$-th iteration, Step 1 aims to update $\mathbf{A}_k$, $\mathbf{B}_k$, $\mathbf{C}_k$, and $\mathbf{D}_k$ with fixed $\mathbf{X}_k$ through\footnote{Two forms of $\mathbf{B}_k$ are further interpreted in \ref{apdx:two_forms_of_B}.
}
\begin{align}
\mathbf{A}_k &= \mathbf{U}^\transpose, \label{eq:update_A} \\
\mathbf{B}_k &= \begin{cases}
[\mathbf{V}, \mathbf{O}_{n \times (m-n)}]^\transpose , & m \geq n , \\
[\mathbf{v}_1,\cdots,\mathbf{v}_m]^\transpose, & m < n, \\
\end{cases} \label{eq:update_B} \\
\mathbf{C}_k &= [\mathbf{u}_1,\cdots,\mathbf{u}_r]^\transpose, \label{eq:update_C} \\
\mathbf{D}_k &= [\mathbf{v}_1,\cdots,\mathbf{v}_r]^\transpose, \label{eq:update_D} 
\end{align}
where $\mathbf{U} \dimension{m \times n}$ and $\mathbf{V} \dimension{n \times n}$ are the left and right orthogonal matrices of $\mathbf{X}_k$'s singular value decomposition, and $r \leq \min(m,n)$ is the number of truncated singular values.

In Step 2, by keeping other variables invariant, $\mathbf{X}_{k+1}$ is optimized via the following problem: 
\begin{equation}	\label{eq:X_minimization}
\begin{aligned}
\min_{\mathbf{X}} \,\ & \trace{\mathbf{A}_k\mathbf{X}\mathbf{B}_k^\transpose} - \trace{\mathbf{C}_k\mathbf{X}\mathbf{D}_k^\transpose} \\
\text{s.t.} \,\ & \, \mathbf{X}_{\Omega} = \mathbf{M}_{\Omega}.
\end{aligned}
\end{equation}

In accordance with common strategies, the alternating direction method of multipliers is naturally adopted to solve \eqref{eq:X_minimization}. After adding an auxiliary variable $\mathbf{W} \dimension{m \times n}$ to relax the constraint, \eqref{eq:X_minimization} can be reformulated as
\begin{equation}	\label{eq:X_minimization_with_W}
\begin{aligned}
\min_{\mathbf{X}} \,\ & \trace{\mathbf{A}_k\mathbf{X}\mathbf{B}_k^\transpose} - \trace{\mathbf{C}_k\mathbf{W}\mathbf{D}_k^\transpose} \\
\text{s.t.} \,\ & \, \mathbf{X} = \mathbf{W}, \ \mathbf{W}_{\Omega} = \mathbf{M}_{\Omega}.
\end{aligned}
\end{equation}
Then, the formulation is converted to an unconstrained augmented Lagrangian function, i.e.,
\begin{align} \label{eq:TNNR_Lagrangian}
\mathcal{L} (\mathbf{X},\mathbf{W},\mathbf{Y}) = &\ \trace{\mathbf{A}_k\mathbf{X}\mathbf{B}_k^\transpose} - \trace{\mathbf{C}_k\mathbf{W}\mathbf{D}_k^\transpose} + \frac{\mu}{2} \norm{\mathbf{X}-\mathbf{W}}_\fro^2 \nonumber \\
& + \trace{\mathbf{Y}^\transpose(\mathbf{X}-\mathbf{W})},
\end{align}
where $\mu > 0$ is a regularizing parameter and $\mathbf{Y} \dimension{m \times n}$ is a Lagrange multiplier matrix.

To achieve the recovery of some missing elements in $\mathbf{X}$ with higher priority and accuracy in each step, partial rows and columns of the equality constraint ($\mathbf{X}-\mathbf{W}$) are allocated with different weights, respectively. As a result, the augmented Lagrangian function \eqref{eq:TNNR_Lagrangian} becomes
\begin{equation}	\label{eq:Weighted_TNNR_Lagrangian}
\begin{aligned}
\mathcal{L} (\mathbf{X},\mathbf{W},\mathbf{Y}) &= \trace{\mathbf{A}_k\mathbf{X}\mathbf{B}_k^\transpose} - \trace{\mathbf{C}_k\mathbf{W}\mathbf{D}_k^\transpose} \\
&  + \frac{\mu}{2} \norm{\mathbf{P}(\mathbf{X}-\mathbf{W})\mathbf{Q}}_\fro^2 + \trace{\mathbf{Y}^\transpose \mathbf{P}(\mathbf{X}-\mathbf{W})\mathbf{Q}},
\end{aligned}
\end{equation}
where the weights are $\mathbf{P} = \text{diag}(\hat{p}_1,\cdots,\hat{p}_m)$, $\{\hat{p}_i\}_{i=1}^m \geq 0$ and $\mathbf{Q} = \text{diag}(\hat{q}_1,\cdots,\hat{q}_n)$, $\{\hat{q}_i\}_{i=1}^n \geq 0$. A large $\hat{p}_i$ ($\hat{q}_i$) leads to the $i$-th row (column) of $\mathbf{X}$ to be restored with high priority and accuracy than the other rows (columns). Based on the number of known elements in different rows and columns, we define
\begin{equation}
\begin{cases}
\hat{p}_i \leq \hat{p}_k , & \text{if } N^{\text{r}}_i \leq N^{\text{r}}_k,\ i,k=1,2,\cdots,m , \\
\hat{q}_j \leq \hat{q}_l , & \text{if } N^{\text{c}}_j \leq N^{\text{c}}_l,\ j,l=1,2,\cdots,n , \\
\end{cases}
\end{equation}
where $N^{\text{r}}_i$ denotes the number of observed elements in the $i$-th row of $\mathbf{X}$, and $N^{\text{c}}_j$ denotes the number of observed elements in the $j$-th column of $\mathbf{X}$.

\subsection{Optimization}


Suppose $\mathbf{X}_t$, $\mathbf{W}_t$, and $\mathbf{Y}_t$ indicate the results of the $t$-th\footnote{Note that we consider the iterative optimization in Step 2 of the TNNR as an inner loop. So let $t$ indicate the number of iterations inside Step 2.} iteration in Step 2. Then, by keeping other variables invariant, $\mathbf{W}_{t+1}$ is updated through
\begin{align}
&\mathbf{W}_{t+1} = \arg\min_{\mathbf{W}} \ \mathcal{L} (\mathbf{X}_t,\mathbf{W},\mathbf{Y}_t) \nonumber \\
&= \arg\min_{\mathbf{W}} \ - \, \trace{\mathbf{C}_k\mathbf{W}\mathbf{D}_k^\transpose} + \frac{\mu_t}{2} \Big\| \mathbf{P}(\mathbf{X}_t-\mathbf{W})\mathbf{Q} + \frac{\mathbf{Y}_t}{\mu_t} \Big\|_\fro^2 \nonumber \\
&= \mathbf{X}_t + \frac{1}{\mu_t} ( \mathbf{P}^{-2} \mathbf{C}_k^\transpose \mathbf{D}_k \mathbf{Q}^{-2} + \mathbf{P}^{-1} \mathbf{Y}_t \mathbf{Q}^{-1} ) . \label{eq:update_W}
\end{align}

Based on the constraint in \eqref{eq:X_minimization_with_W}, the values of observed elements should remain unchanged during the updating process. Hereby, we obtain
\begin{gather}
\mathbf{W}_{t+1} = (\mathbf{W}_{t+1})_{\Omega^{\text{c}}} + \mathbf{M}_{\Omega}, \\
(\mathbf{W}_{\Omega^\text{c}})_{ij} = \begin{cases}
\mathbf{W}_{ij}, & (i,j) \in \Omega^\text{c}, \\
0, & (i,j) \in \Omega \, ,
\end{cases}
\end{gather}
where $\Omega^{\text{c}}$ includes the indices of missing elements in data.

By fixing $\mathbf{W}_{t+1}$ and $\mathbf{Y}_t$, $\mathbf{X}_{t+1}$ is computed via
\begin{align}
&\mathbf{X}_{t+1} = \arg\min_{\mathbf{X}} \ \mathcal{L} (\mathbf{X},\mathbf{W}_{t+1},\mathbf{Y}_t) \nonumber \\
&= \arg\min_{\mathbf{X}} \ \trace{\mathbf{A}_k\mathbf{X}\mathbf{B}_k^\transpose} + \frac{\mu_t}{2} \Big\|\mathbf{P}(\mathbf{X}-\mathbf{W}_{t+1})\mathbf{Q} + \frac{\mathbf{Y}_t}{\mu_t} \Big\|_\fro^2 \nonumber \\
&= \mathbf{W}_{t+1} - \frac{1}{\mu_t} ( \mathbf{P}^{-2} \mathbf{A}_k^\transpose \mathbf{B}_k \mathbf{Q}^{-2} + \mathbf{P}^{-1} \mathbf{Y}_t \mathbf{Q}^{-1} ). \label{eq:updata_X}
\end{align}

Subsequently, $\mathbf{Y}_{t+1}$ is updated directly as follows: 
\begin{equation} \label{eq:update_Y}
\mathbf{Y}_{t+1} = \mathbf{Y}_t + \beta_t \mathbf{P}(\mathbf{X}_{t+1}-\mathbf{W}_{t+1})\mathbf{Q} \, ,
\end{equation}
where $\beta_t > 0$ is set as a monotonically increasing sequence, which is usually beneficial for convergence.

Although three terms are solved in closed form solutions, they still require a large number of iterations to converge in practice. In the light of intrinsic correlations in \eqref{eq:update_W}--\eqref{eq:update_Y}, we derive a concise gradient descent manner to efficiently update $\mathbf{X}_k$ without substantial iterative steps, as revealed in Theorem~\ref{thm:X_update}. 

\begin{theorem} \label{thm:X_update}
	If $0 < \mu_t < \mu_{t+1}$, $\forall\, t=1,2,\cdots, N$ and $\frac{1}{\alpha_k} = \sum_{t=1}^{N-1} \frac{1}{\mu_t}$. Derived from \eqref{eq:update_W}--\eqref{eq:update_Y}, the updating step of $\mathbf{X}_k$ is concisely formulated as one-step computation:
		\begin{gather}
		\mathbf{X}^* = \mathbf{X}_k - \frac{1}{\alpha_k} \mathbf{P}^{-2} (\mathbf{A}_k^\emph{\transpose} \mathbf{B}_k - \mathbf{C}_k^\emph{\transpose} \mathbf{D}_k) \mathbf{Q}^{-2}, \\
		\mathbf{X}_{k+1} = (\mathbf{X}^*)_{\Omega^\text{\emph{c}}} + \mathbf{M}_{\Omega},
		\end{gather}
		where $\frac{1}{\alpha_k} > 0$ stands for a step size.
\end{theorem}

Proof of Theorem~\ref{thm:X_update} is offered in \ref{apdx:proof_X_update}.

For brief notations, we define two weight matrices as 
\begin{align}
\mathcal{P} &= \mathbf{P}^{-2} = \text{diag}(p_1,\cdots,p_m), \label{eq:P} \\
\mathcal{Q} &= \mathbf{Q}^{-2} = \text{diag}(q_1,\cdots,q_n), \label{eq:Q} 
\end{align}
where $\{p_i\}_{i=1}^m \geq 0$ and $\{q_j\}_{j=1}^n \geq 0$ are precisely determined based on the number of observed elements in each row and column, respectively, as follows: 
\begin{align}
p_i &= \exp \left(-\theta_1 \left( \frac{N^{\text{r}}_i}{n} - 1 \right) \right)-1,\ i=1,2,\cdots,m, \label{eq:p_expo} \\
q_j &= \exp \left(-\theta_2 \left( \frac{N^{\text{c}}_j}{m} - 1 \right) \right)-1,\ j=1,2,\cdots,n, \label{eq:q_expo}
\end{align}
where $\theta_1$ and $\theta_2$ scale the weights. Clearly, we have $p_i \leq p_j$ if $N^{\text{r}}_i \geq N^{\text{r}}_j$ and $q_i \leq q_j$ if $N^{\text{c}}_i \geq N^{\text{c}}_j$, i.e., the row (column) with more observed elements is assigned with a smaller value of weight, according to \eqref{eq:p_expo} and \eqref{eq:q_expo}. Note that $p_i = 0$ ($q_j = 0$) implies that the $i$-th row ($j$-th column) has no element lost.

Moreover, we discover the property that
\begin{align}
\mathbf{A}_k &= \mathbf{U}_k^\transpose = [\mathbf{u}_1,\cdots,\mathbf{u}_r,\cdots,\mathbf{u}_m]^\transpose = [\mathbf{C}_k^\transpose, \mathbf{u}_{r+1}, \cdots, \mathbf{u}_m]^\transpose \nonumber \\
&= [\mathbf{C}_k^\transpose,\mathbf{\varPhi}_k^\transpose]^\transpose, \\
\mathbf{B}_k &= [\mathbf{v}_1,\cdots,\mathbf{v}_r,\cdots,\mathbf{v}_m]^\transpose = [\mathbf{D}_k^\transpose, \mathbf{v}_{r+1}, \cdots, \mathbf{v}_m]^\transpose \nonumber \\
&= [\mathbf{D}_k^\transpose,\mathbf{\varLambda}_k^\transpose]^\transpose.
\end{align}
With the above property, we further infer the rule that
\begin{equation}
\mathbf{A}_k^\transpose \mathbf{B}_k - \mathbf{C}_k^\transpose \mathbf{D}_k = [\mathbf{C}_k^\transpose, \mathbf{\varPhi}_k^\transpose] \begin{bmatrix}
\mathbf{D}_k \\
\mathbf{\varLambda}_k \\
\end{bmatrix}
- \mathbf{C}_k^\transpose \mathbf{D}_k = \mathbf{\varPhi}_k^\transpose \mathbf{\varLambda}_k, \label{eq:simplify_AB_CD}
\end{equation}
where
\begin{equation}
\mathbf{\varPhi}_k = [\mathbf{u}_{r+1}, \cdots, \mathbf{u}_m]^\transpose, \ \mathbf{\varLambda}_k = [\mathbf{v}_{r+1}, \cdots, \mathbf{v}_m]^\transpose . \label{eq:Psi_and_Lambda} \\
\end{equation}

Hence, by means of Theorem~\ref{thm:X_update}, \eqref{eq:P}--\eqref{eq:q_expo}, and \eqref{eq:simplify_AB_CD}, $\mathbf{X}_{k+1}$ can be calculated efficiently (without a number of iterations) in a simple form of the gradient descend method:
\begin{gather}
\mathbf{X}^* = \mathbf{X}_k - \frac{1}{\alpha_k} \mathcal{P} \mathbf{\varPhi}_k^\transpose \mathbf{\varLambda}_k \mathcal{Q}, \label{eq:X_gradient} \\
\mathbf{X}_{k+1} = (\mathbf{X}^*)_{\Omega^\text{c}} + \mathbf{M}_{\Omega}. \label{eq:X_known}
\end{gather}
On the basis of \eqref{eq:X_gradient} and \eqref{eq:X_known}, we find that they are equivalent to the solution by the gradient descent search of function $\trace{\mathbf{\varPhi}_k\mathcal{P}\mathbf{X}\mathcal{Q}\mathbf{\varLambda}_k^\transpose}$.

Due to the merit of the one-step gradient descent manner instead of computing $\mathbf{X}_k$ iteratively in Step 2, in summary, we can convert \eqref{eq:rewritten_TNNR_objective} to the following formulation:
\begin{equation} \label{eq:weighted_TNNR_minimization}
\min_{\mathbf{X}} \,\  \trace{\mathbf{\varPhi}\mathcal{P}\mathbf{X}\mathcal{Q}\mathbf{\varLambda}^\transpose} \quad
\text{s.t.} \,\ \mathbf{X}_{\Omega} = \mathbf{M}_{\Omega},
\end{equation}
where $	\mathcal{P} = \text{diag}(p_1,\cdots,p_m)$, $\mathcal{Q} = \text{diag}(q_1,\cdots,q_n)$, $\mathbf{\varPhi}_k = [\mathbf{u}_{r+1}, \cdots, \mathbf{u}_m]^\transpose$, and $\mathbf{\varLambda}_k = [\mathbf{v}_{r+1}, \cdots, \mathbf{v}_m]^\transpose$. $\mathbf{U}$ and $\mathbf{V}$ denote the left and right orthogonal matrices generated by $\mathbf{X}$'s SVD. $r \leq \min(m,n)$ is the number of truncated singular values. 

Note that $r =m$ indicates $\mathbf{\varPhi}_k = \mathbf{O}$ and $\mathbf{\varLambda}_k=\mathbf{O}$, which implies the gradient vanishes. Nevertheless, we choose $r \ll m$ empirically, so the undesirable situation will not occur, which conforms the fact that visual data (e.g.~a real image) has low-rank structure ubiquitously. In other word, the actual rank $r$ is much smaller than the dimension of an image.

With an initial $\mathbf{X}_1 = \mathbf{M}_{\Omega}$, the optimization \eqref{eq:weighted_TNNR_minimization} is solved by a gradient descent step as follows:
\begin{gather}
\mathbf{X}_{k+1} = \mathbf{X}_k - \frac{1}{\alpha_k} \mathcal{P} \mathbf{\varPhi}_k^\transpose \mathbf{\varLambda}_k \mathcal{Q}, \label{eq:update_X_simple_gradient} \\
\mathbf{X}_{k+1} = (\mathbf{X}_{k+1})_{\Omega^\text{c}} + \mathbf{M}_{\Omega},
\end{gather}
where $\frac{1}{\alpha_k}$ is a diminishing step size that satisfies 
\begin{equation}
\alpha_{k+1} = \rho \, \alpha_k,
\end{equation}
where $\rho > 1$ is a constant. Furthermore, the required number of iterations conforms to Theorem~\ref{thm:alpha_proof}.

\begin{theorem} \label{thm:alpha_proof}
	If $0 < \alpha_k < \alpha_{k+1}$, $\lim_{k \rightarrow \infty} \frac{1}{\alpha_k}=0$, $\varepsilon$ is a stopping tolerance, and the result of our DW-TNNR method converges, i.e.~$\norm{\mathbf{X}_{N+1}-\mathbf{X}_N}_\text{\emph{F}} \leq \varepsilon$, then the number of iterations satisfies 
	\begin{equation}
	N \geq 1 + \frac{\ln \gamma - \ln (\alpha_1 \varepsilon)}{\ln \rho} , \ \gamma = \norm{\mathcal{P}}_\text{\emph{F}} \, (\sqrt{m} + \sqrt{r}) \, \norm{\mathcal{Q}}_\text{\emph{F}}, \nonumber
	\end{equation}
	where $N$ is the number of iterations, $m$ is the number of rows in $\mathbf{X}$, and $r$ indicates the number of truncated singular values.
\end{theorem}

Proof of Theorem~\ref{thm:alpha_proof} is provided in \ref{apdx:proof_alpha}.

The compact optimization procedures are summarized in Algorithm~\ref{alg:DW-TNNR}. 

\begin{algorithm}[t] \small
	\caption{Double weighted truncated nuclear norm regularization method}
	\label{alg:DW-TNNR}
	\begin{algorithmic}[1]
		\REQUIRE $\mathbf{M}$, incomplete matrix; $\Omega$, the index set of known elements; $\Omega^\text{c}$, the index set of lost elements.
		\ENSURE $\mathbf{X}_1=\mathbf{M}_\Omega$, $k=1$, $K = 200$, $\alpha_1$, $\rho$, $\varepsilon$; \\
		$\mathcal{P}$ and $\mathcal{Q}$, two weight matrices for the row and column of $\mathbf{X}$. \\
		\WHILE {$\Delta \geq \varepsilon$ and $k \leq K$}
		\STATE \begin{equation}
		[\mathbf{U}_k,\Sigma_k,\mathbf{V}_k]=\text{svd}(\mathbf{X}_k). \nonumber
		\end{equation} \vspace{-2ex}
		\STATE Update  $\mathbf{\varPhi}_k$ and $\mathbf{\varLambda}_k$ through
		\begin{equation}
		\mathbf{\varPhi}_k = [\mathbf{u}_{r+1}, \cdots, \mathbf{u}_m]^\transpose , \ \mathbf{\varLambda}_k = [\mathbf{v}_{r+1}, \cdots, \mathbf{v}_m]^\transpose . \nonumber	
		\end{equation} \vspace{-2ex}
		\STATE Compute $\mathbf{X}_{k+1}$ using the gradient descent method:
		\begin{align}
		\mathbf{X}_{k+1} &= \mathbf{X}_k - \frac{1}{\alpha_k} \mathcal{P} \mathbf{\varPhi}_k^\transpose \mathbf{\varLambda}_k \mathcal{Q}, \nonumber \\
		\mathbf{X}_{k+1} &= (\mathbf{X}_{k+1})_{\Omega^\text{c}} + \mathbf{M}_{\Omega}. \nonumber
		\end{align} \vspace{-2ex}
		\STATE Update the step size: 
		\begin{equation}
		\alpha_{k+1} = \rho \, \alpha_k . \nonumber
		\end{equation} \vspace{-2ex}
		\STATE Save the intermediate result: $\mathbf{X}_\text{rec}=\mathbf{X}_{k+1}$.
		\STATE Compute the relative variation: 
		\begin{equation}
		\Delta = \norm{\mathbf{X}_{k+1}-\mathbf{X}_k}_\fro / \norm{\mathbf{M}}_\fro. \nonumber
		\end{equation} \vspace{-2ex}
		\STATE $k=k+1$.
		\ENDWHILE
		\RETURN the optimal recovered image $\mathbf{X}_\text{rec}$.
	\end{algorithmic}
\end{algorithm}

\section{Experiment} \label{sec:experiment}

To demonstrate the effectiveness of our proposed method, substantial experiments are performed with several compared approaches. They are as follows:
\begin{enumerate}
	\item TNNR \cite{Hu2013-Accurate};
	\item IRNN-L$_\text{p}$ \cite{Lu2016-Nonconvex};
	\item PSSV \cite{Oh2016-Partial};
	\item Joint Schatten-$p$ norm (Joint S$_\text{p}$) \cite{Nie2015-Joint};
	\item DW-TNNR [ours];
\end{enumerate} 
All experiments are executed in Matlab R2015b based on Windows 10, equipped with an Intel Core i7 CPU @ 2.60 GHz and 12 GB Memory.

According to Theorems~\ref{thm:X_update} and \ref{thm:alpha_proof}, the most important parameters of the DW-TNNR are weight matrices $\mathcal{P}$ and $\mathcal{Q}$, which are decided by \eqref{eq:p_expo} and \eqref{eq:q_expo}, respectively. Without prior knowledge to the number of truncated singular values, the parameter $r$ is examined from range [1, 20] to select an optimal value for each case.

The maximum number of iterations $K$ is 200 for the DW-TNNR and TNNR methods, but it is fixed to 500 for the other approaches. All algorithms are considered to have converged if $\norm{\mathbf{X}_{k+1}-\mathbf{X}_k}_\fro / \norm{\mathbf{M}}_\fro < \varepsilon$. We set $\varepsilon = 10^{-4}$ on the real visual data for all compared approaches. All adjustable parameters of the compared methods are tuned to be optimal. We report the best results in this section. For fair comparison, each result is averaged over 10 separate runs.

The overall error of reconstruction (Erec) and the peak signal-to-noise ratio (PSNR) are frequently adopted metrics in matrix completion for evaluating the performances of various algorithms. They are defined as follows: 
\begin{align}
\text{Erec} &= \norm{(\mathbf{X}_\text{rec} - \mathbf{M})_{\Omega^\text{c}}}_\fro, \\
\text{SE} &= \text{Erec}_\text{R}^2 + \text{Erec}_\text{G}^2 + \text{Erec}_\text{B}^2, \\
\text{MSE} &= \frac{\text{SE}}{T} = \frac{\text{SE}}{T_\text{R} + T_\text{G} + T_\text{B}}, \\
\text{PSNR} &= 10 \times \log_{10} \left( \frac{255^2}{\text{MSE}} \right) ,
\end{align}
where $T$ is the total number of lost elements, and \{R, G, B\} denotes three channels of a color image. For the intuitive description on the weights corresponding to missing parts, we define the following matrix as a visualization to the values of weights, which is given by
\begin{equation}
\mathcal{W} = (\mathcal{P} \cdot \mathbf{1}_{m \times n} \cdot \mathcal{Q})_{\Omega^\text{c}},
\end{equation}
where $\mathbf{1}_{m \times n}$ implies an $m \times n$ matrix, all elements of which are equal to 1.

\subsection{Effects of $\mathcal{P}$ and $\mathcal{Q}$ of DW-TNNR}

The weight matrices $\mathcal{P}$ and $\mathcal{Q}$ are significant for the DW-TNNR algorithm, which control the priority of convergence to each row and each column of $\mathbf{X} \in \mathbb{R}^{m \times n}$ in optimization. 

\begin{figure*}[t]
	\centering
	\subfloat[]{\includegraphics[width=0.19\textwidth]{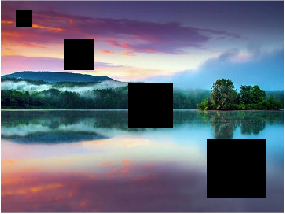}%
		\label{fig:4_blocks_incomplete}
	}
	\hfil
	\subfloat[]{\includegraphics[width=0.19\textwidth]{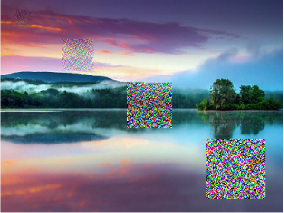}%
		\label{fig:4_blocks_iter_10}
	}
	\hfil
	\subfloat[]{\includegraphics[width=0.19\textwidth]{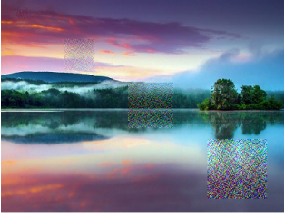}%
		\label{fig:4_blocks_iter_20}
	}
	\hfil
	\subfloat[]{\includegraphics[width=0.19\textwidth]{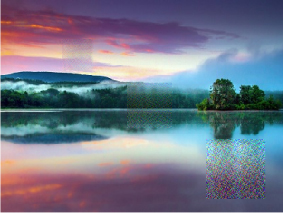}%
		\label{fig:4_blocks_iter_30}
	}
	\hfil
	\subfloat[]{\includegraphics[width=0.19\textwidth]{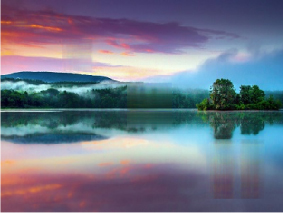}%
		\label{fig:4_blocks_iter_result}
	}
	\caption{Process of DW-TNNR on an image with four missing blocks: (a) incomplete image, (b) step 10, (c) step 20, (d) step 30, (e) recovered image}
	\label{fig:process_4_blocks}
\end{figure*}

An image with a size of $400 \times 300$ is corrupted with four different sizes of missing blocks (Fig.~\ref{fig:4_blocks_incomplete}). It is recovered by the DW-TNNR approach with the parameters $\theta_1 = \theta_2 = 1.2$, $\alpha_1 = 10^{-4}$, and $\rho = 1.2$. 

\begin{figure}[t]
	\centering
	\includegraphics[scale=0.60]{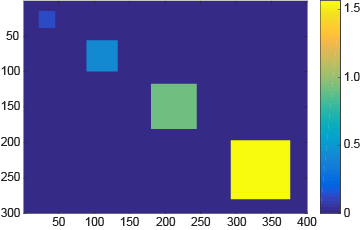}
	\caption{Weights corresponding to four missing blocks in Fig.~\ref{fig:4_blocks_incomplete}}
	\label{fig:weights_4_blocks}
\end{figure}

Figs.~\ref{fig:4_blocks_iter_10}--\ref{fig:4_blocks_iter_result} illustrate that the smaller blocks are recovered prior than the bigger ones. For clarification, Fig.~\ref{fig:weights_4_blocks} depicts the weights applied in Fig.~\ref{fig:process_4_blocks}. The smaller blocks are distributed with the smaller values, since they are commonly easier to be restored than the others. Likewise, the bigger blocks require the larger values of weights to ensure the accuracy. Therefore, the entire process is accelerated by the sequential recovery fashion.

\begin{figure*}[t]
	\centering
	\subfloat[]{\includegraphics[scale=0.36]{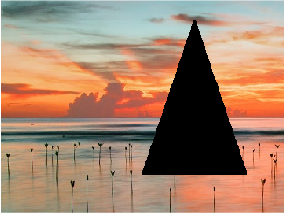}%
		\label{fig:triangle_incomplete}
	}
	\hfil
	\subfloat[]{\includegraphics[scale=0.36]{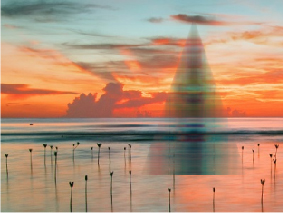}%
		\label{fig:triangle_recover_bad}
	}
	\hfil
	\subfloat[]{\includegraphics[scale=0.36]{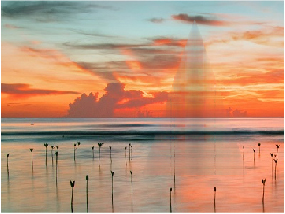}%
		\label{fig:triangle_recover_good}
	}
	\hfil
	\subfloat[]{\includegraphics[scale=0.32]{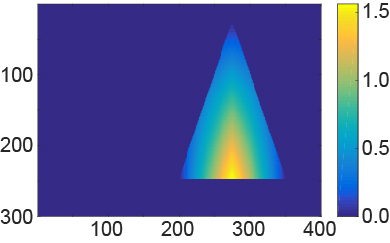}%
		\label{fig:triangle_weight}
	}
	\\[-2.0mm]
	\subfloat[]{\includegraphics[scale=0.36]{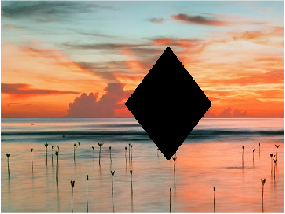}%
		\label{fig:diamond_incomplete}
	}
	\hfil
	\subfloat[]{\includegraphics[scale=0.36]{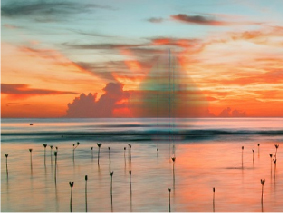}%
		\label{fig:diamond_recover_bad}
	}
	\hfil
	\subfloat[]{\includegraphics[scale=0.36]{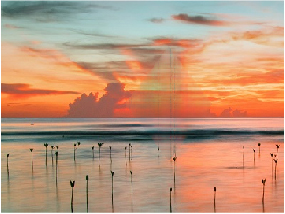}%
		\label{fig:diamond_recover_good}
	}
	\hfil
	\subfloat[]{\includegraphics[scale=0.32]{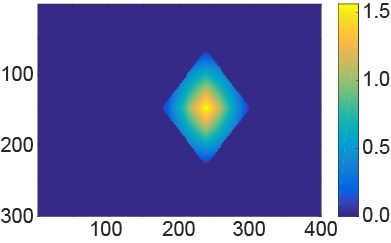}%
		\label{fig:diamond_weight}
	}
	\caption{Recovery of two kinds of incomplete images by DW-TNNR: (a) triangular block; (b) no weight, PSNR = 15.39; (c) double weighted, PSNR = 19.32; (d) visualized weights; (e) diamond block; (f) no weight, PSNR = 15.64; (g) double weighted, PSNR = 19.67; (h) visualized weights.}
	\label{fig:result_triangle_recovery}
\end{figure*}

In Fig.~\ref{fig:result_triangle_recovery}, two kinds of missing blocks are illustrated in Figs.~\ref{fig:triangle_incomplete} and \ref{fig:diamond_incomplete}. In such scenarios, the number of unknown elements in data varies successively in both row and column dimensions. Compared with no weight cases ($\mathcal{P}=\mathbf{I}$, $\mathcal{Q}=\mathbf{I}$), as shown in Figs.~\ref{fig:triangle_recover_bad}~and~\ref{fig:diamond_recover_bad}, the DW-TNNR method performs better on the triangular block (Fig.~\ref{fig:triangle_recover_good}) and the diamond block (Fig.~\ref{fig:diamond_recover_good}). Here, we still keep $\theta_1=\theta_2=1.2$, $\alpha_1 = 10^{-4}$, and $\rho = 1.2$. Besides, we depict the weights in two cases, as shown in Figs.~\ref{fig:triangle_weight}~and~\ref{fig:diamond_weight}. It implies that the row (column) with more unknown elements is allocated with a larger value of weight. It is observable that the recovery results with assigned weights (Figs.~\ref{fig:triangle_recover_good} and \ref{fig:diamond_recover_good}) are better than those with no weight (Figs.~\ref{fig:triangle_recover_bad} and \ref{fig:diamond_recover_bad}), respectively. Thus, it verifies that the effectiveness of the weighting scheme in our algorithm.

\begin{figure}[t]
	\centering
	\includegraphics[scale=0.40]{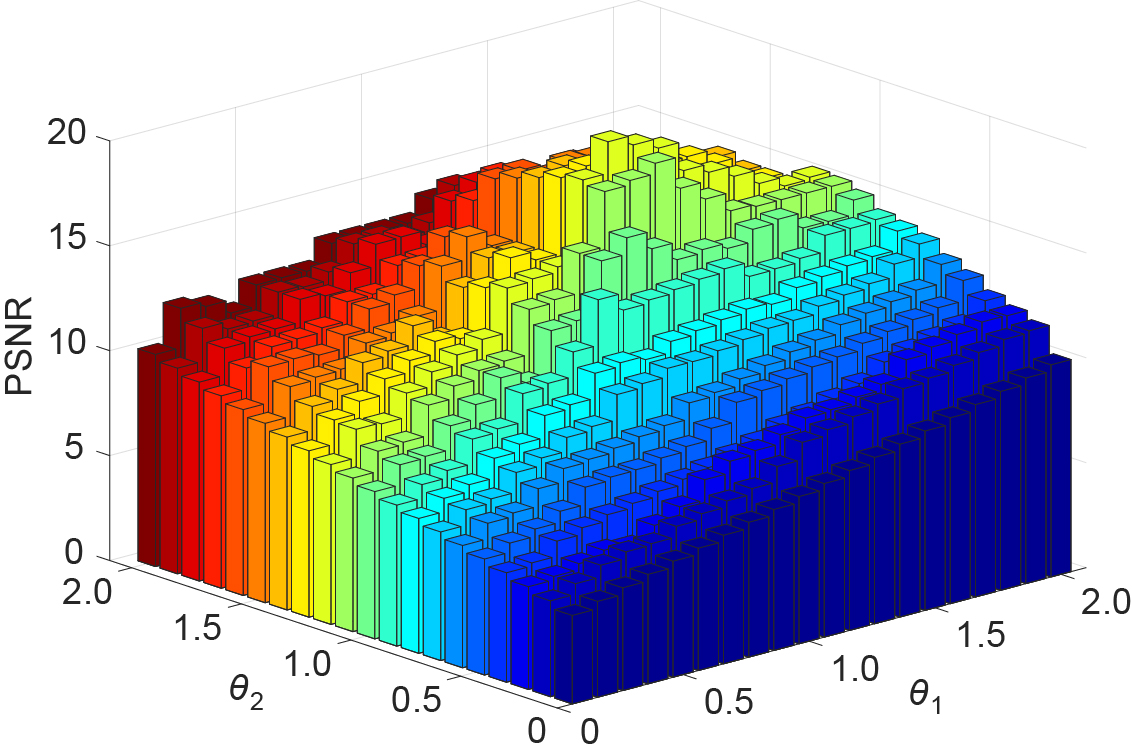}
	\caption{Performance of the DW-TNNR under different values of $\theta_1$ and $\theta_2$ on Fig.~\ref{fig:triangle_incomplete}}
	\label{fig:varying_theta_comparison}
\end{figure}

Then, the effects of $\theta_1$ and $\theta_2$ are individually investigated. We conduct extra experiments by changing the value of $\theta_1$ or $\theta_2$ while fixing another one. The incomplete image Fig.~\ref{fig:triangle_incomplete} is used and the PSNR is reported in Fig.~\ref{fig:varying_theta_comparison}. Let $\alpha_1 = 10^{-4}$ and $\rho = 1.2$. We discover that the best result occurs merely when $\theta_1$ and $\theta_2$ are equal to 1.2. However, if $\theta_1$ or $\theta_2$ is too small or too large, the PSNR becomes worse. Since a smaller value of $\theta_1$ ($\theta_2$) has tiny effect and a larger value of $\theta_1$ ($\theta_2$) impairs the recovery of our method. Note that $\theta_1$ and $\theta_2$ control the range of weights in the row and column dimension, respectively. It is natural that they may have identical impact on the recovery by the DW-TNNR. Hence, we decide that $\theta_1=\theta_2 = 1.2$ for our method and apply it throughout the experiments.

\subsection{Robustness to $r$ of DW-TNNR}

\begin{figure}[t]
	\centering
	\begin{tabular}{c}
		\includegraphics[scale=0.32]{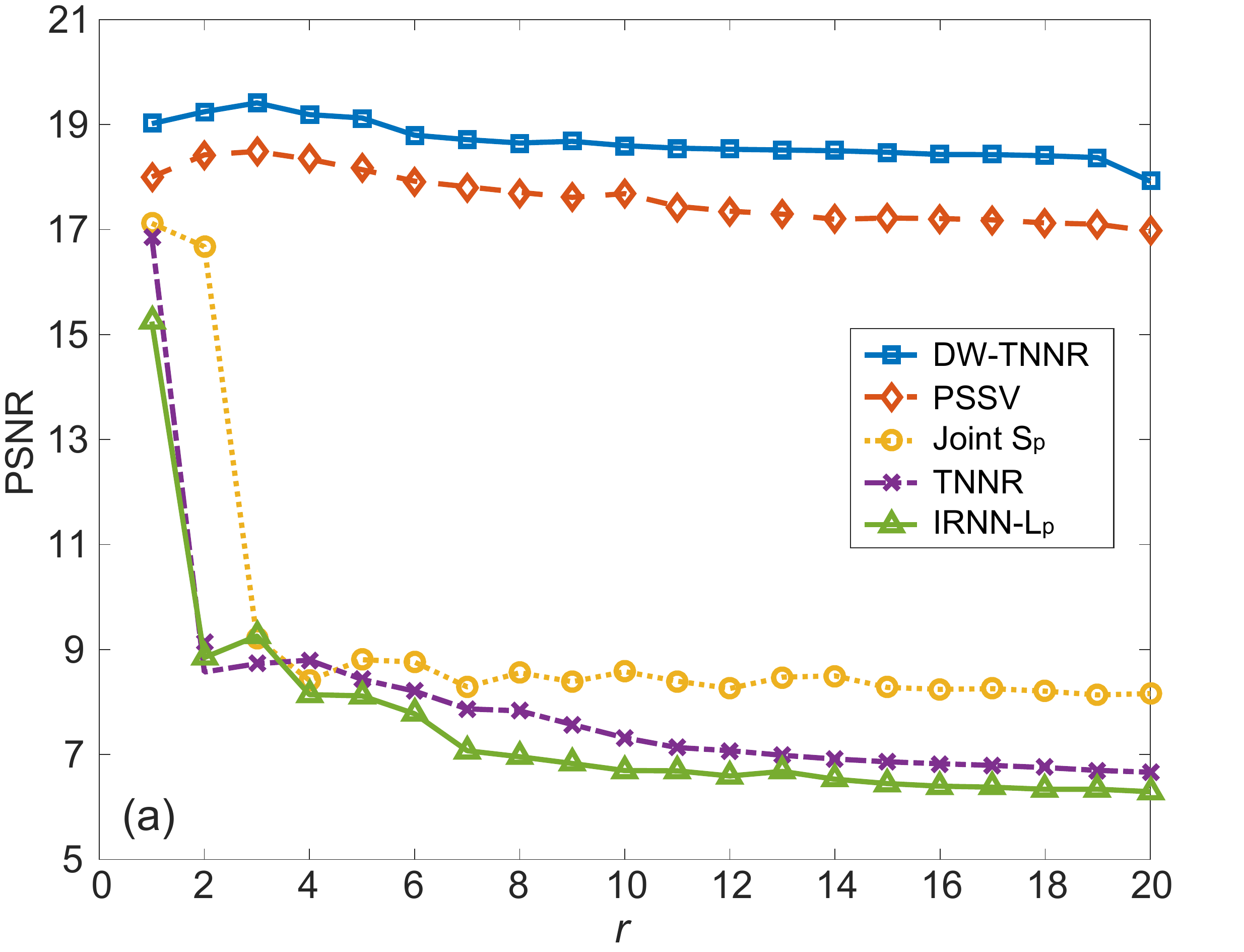}%
		\\
		\includegraphics[scale=0.32]{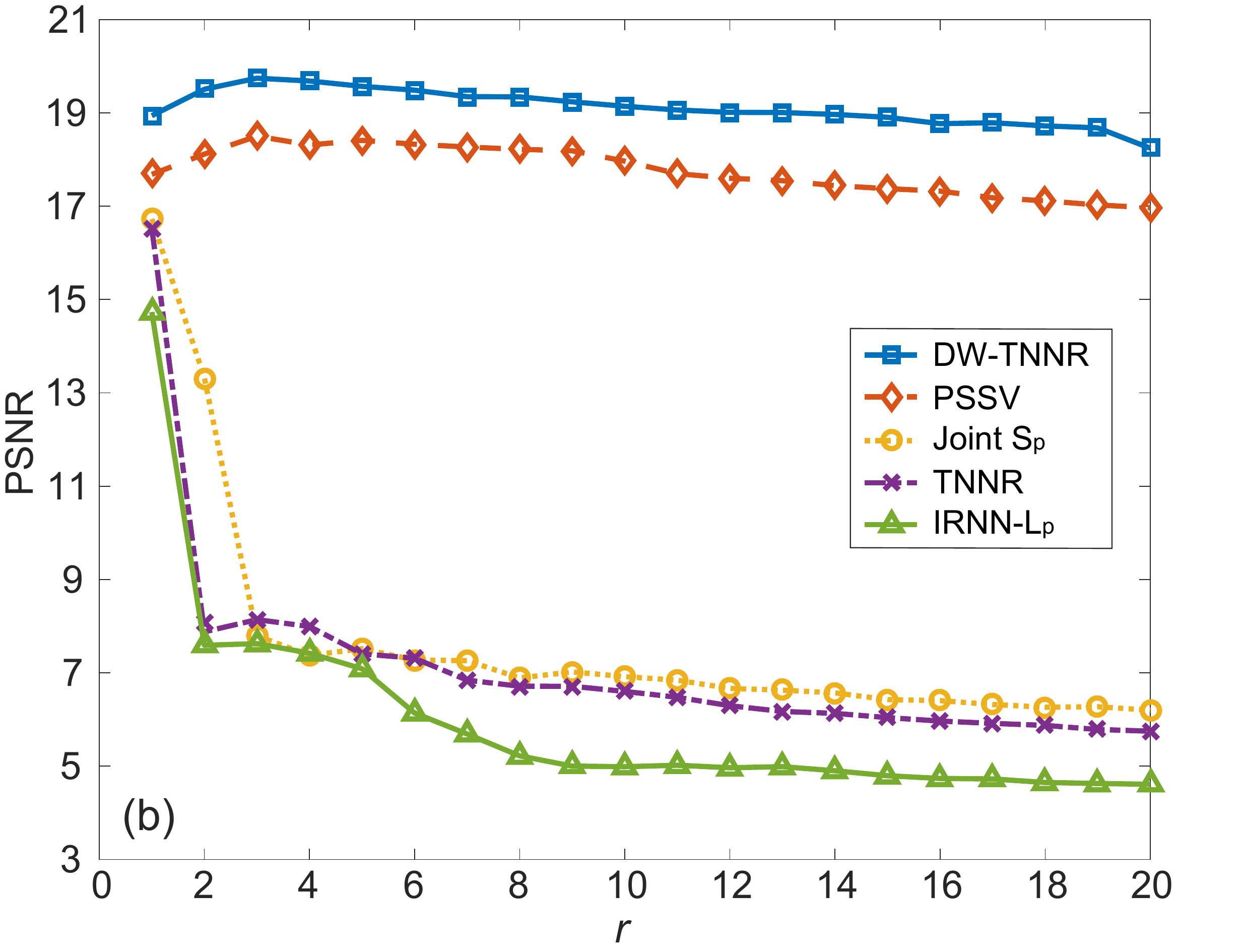}%
	\end{tabular}
	\caption{Comparison of robustness to the number of truncated singular values by the DW-TNNR, PSSV, Joint S$_\text{p}$, TNNR, and IRNN-L$_\text{p}$ methods: (a) with triangular block in Fig.~\ref{fig:triangle_incomplete}; (b) with diamond block in Fig.~\ref{fig:diamond_incomplete}.}
	\label{fig:rank_robust_comparison}
\end{figure}

To validate $r$ is insensitive in our approach, we devise an experiment to verify the robustness to the number of truncated singular values of our DW-TNNR in Figs.~\ref{fig:triangle_incomplete} and \ref{fig:diamond_incomplete}, compared with the PSSV, Joint S$_\text{p}$, TNNR, and IRNN-L$_\text{p}$ methods. Let $\theta_1 = \theta_2 = 1.2$, $\alpha_1 = 10^{-4}$, $\rho = 1.2$, and $r \in [1,20]$. Comparison results are shown in Figs.~\ref{fig:rank_robust_comparison}a and \ref{fig:rank_robust_comparison}b. Clearly, the DW-TNNR and PSSV both have satisfactory robustness to $r$, both of which are relatively steady compared with three other approaches. In addition, the DW-TNNR is slightly superior in PSNR than the PSSV and achieves the optimal performance when $r = 3$. Except for the acceptable recovery of the Joint S$_\text{p}$ when $r=1,2$, its performance deteriorates sharply when $r \geq 3$ and it obtains poor (but quite stable) results in the cases of $r \in [3,20]$. Similarly, the TNNR and IRNN-L$_\text{p}$ have good performances only when $r=1$, and fail to deal with two kinds of incomplete images in the cases of $r \in [2, 20]$. In a word, the DW-TNNR approach can recover a deficient image effectively under various numbers of truncated singular values. Therefore, we do not focus on the specific number of parameter $r$ but set the range $[1,20]$ to select the best result in our experiments.

\subsection{Real Visual Data}

Matrix completion remains a valuable task for image processing, since some images may be corrupted due to noises or occlusions in real-world scenarios. Because a color image has three channels, we deal with each one separately and integrate them eventually for visualization.

\begin{figure}[t]
	\centering
	\includegraphics[width=0.47\textwidth]{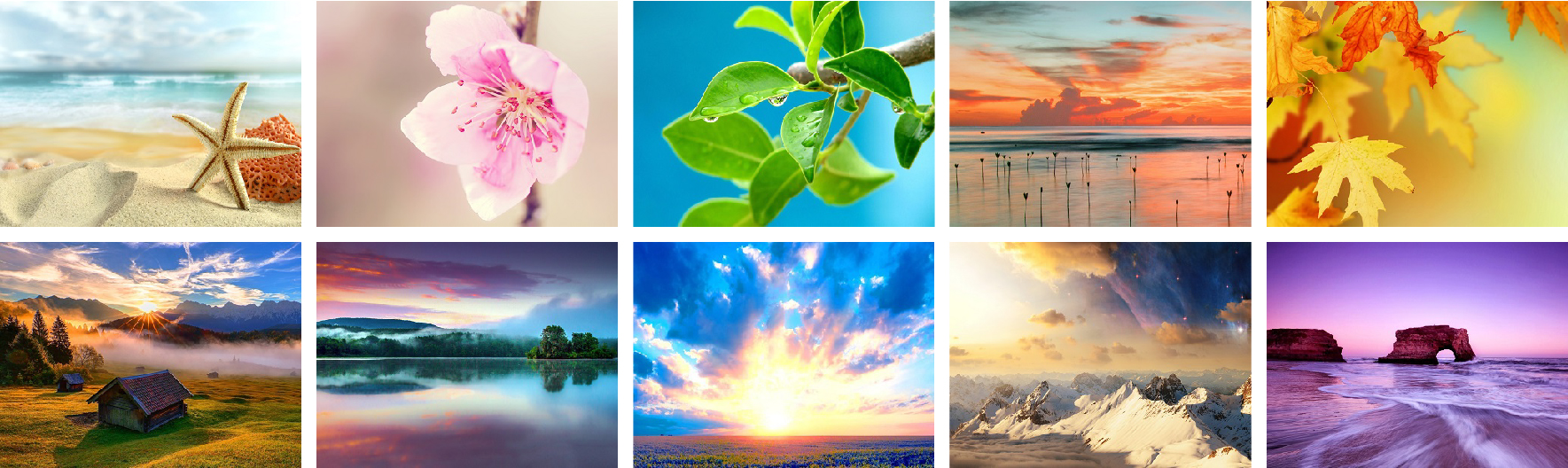}
	\caption{Ten images for experiments used in Section~\ref{sec:experiment}}
	\label{fig:ten_images}
\end{figure}

PSNR is adopted to evaluate the quality of image recovery generated by different algorithms. We use 10 images, as shown in Fig.~\ref{fig:ten_images}, all of which are $400 \times 300$ and will be contaminated by the two types of occlusions: random loss and text block.

\begin{figure}[t]
	\centering
	\subfloat[]{\includegraphics[width=0.15\textwidth]{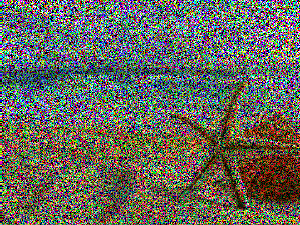}%
		\label{fig:loss_incomplete}
	}
	\subfloat[]{\includegraphics[width=0.15\textwidth]{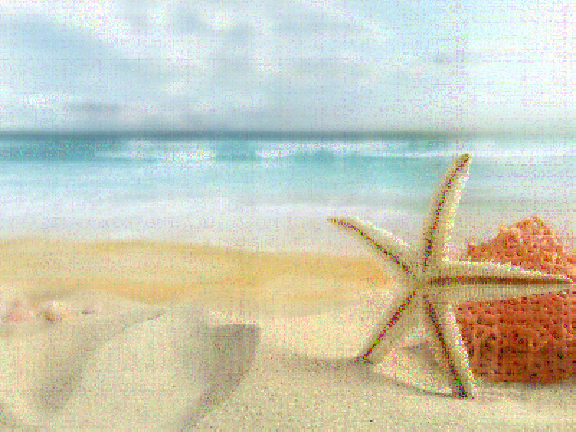}%
		\label{fig:loss_TNNR}
	}
	\subfloat[]{\includegraphics[width=0.15\textwidth]{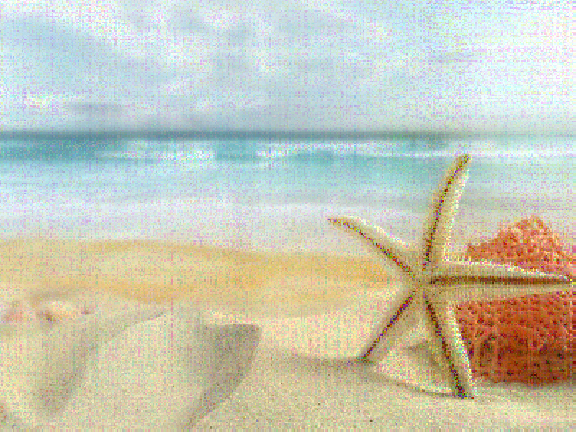}%
		\label{fig:loss_IRNN}
	}
	\\[-2.0mm]
	\subfloat[]{\includegraphics[width=0.15\textwidth]{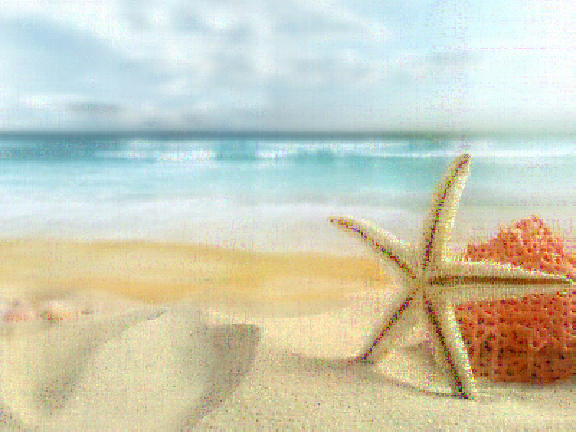}%
		\label{fig:loss_PSSV}
	}
	\subfloat[]{\includegraphics[width=0.15\textwidth]{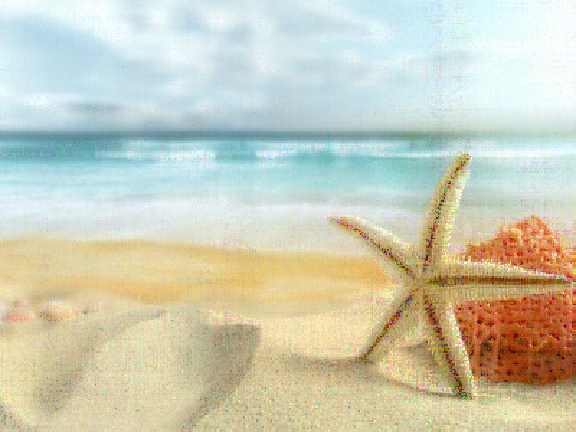}%
		\label{fig:loss_Joint_Sp}
	}
	\subfloat[]{\includegraphics[width=0.15\textwidth]{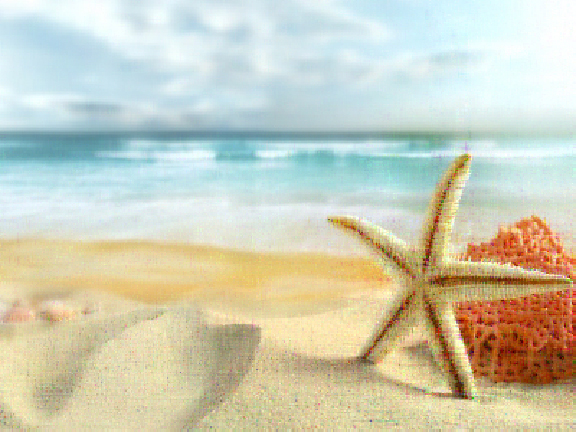}%
		\label{fig:loss_DW_TNNR}
	}
	\caption{Recovered results of the 1st image in Fig.~\ref{fig:ten_images} with random loss (missing ratio = 0.5) by five methods: (a) incomplete image, (b) TNNR, (c) IRNN-L$_\text{p}$, (d) PSSV, (e) Joint S$_\text{p}$, (f) DW-TNNR.}
	\label{fig:visual_loss_images}
\end{figure}

1) \emph{Random loss.} The experiments are executed to recover 10 deficient images whose partial elements are randomly lost (Fig.~\ref{fig:visual_loss_images}a). The missing ratio is fixed to 0.5 for all cases. We set $\theta_1 = \theta_2 = 1.2$, $\alpha_1 = 10^{-4}$, and $\rho = 1.2$ as mentioned earlier. Under this configurations, our DW-TNNR method compares to the TNNR, IRNN-L$_\text{p}$, PSSV, and Joint S$_\text{p}$ approaches. Here, the accuracy of recovery, the iteration number, and the restored image are used as the metrics to evaluate the performances of five approaches. They are demonstrated in Table~\ref{tab:loss_PSNR} and Fig.~\ref{fig:visual_loss_images}, respectively.

In Table~\ref{tab:loss_PSNR}, it is obvious that the PSNR of the DW-TNNR always exceeds those of other approaches. The PSSV and Joint S$_\text{p}$ methods perform close to the our method and much better than the TNNR approach in PSNR on all images. Clearly, the IRNN-L$_\text{p}$ method has the worst recovery results. Furthermore, our DW-TNNR approach runs highly stable on all images and entails about 46 iterations to converge. The iteration number of Joint S$_\text{p}$ method is nearly twice than ours. The IRNN-L$_\text{p}$ and PSSV methods both require more than 100 iterations to converge. Although the TNNR method is slightly better than the IRNN-L$_\text{p}$ method in PSNR, it consumes more than twice iteration numbers to converge. It validates that the DW-TNNR method is highly stable and efficient to recover the incomplete images with random loss.

We only choose the 1st image in Fig.~\ref{fig:ten_images} as an example due to the space limit. This image with 50\% random element loss is shown in Fig.~\ref{fig:loss_incomplete}. Apparently, our DW-TNNR method recovers the 1st image successfully (Fig.~\ref{fig:loss_DW_TNNR}). Some specifics are much clearer visually than the results of the TNNR and IRNN-L$_\text{p}$ methods (Figs.~\ref{fig:loss_TNNR} and \ref{fig:loss_IRNN}) and are a bit better than those of the PSSV and Joint S$_\text{p}$ approaches (Figs.~\ref{fig:loss_PSSV} and \ref{fig:loss_Joint_Sp}). It indicates that the DW-TNNR method is more effective to restore images with random loss.

\begin{table}[t] \small 
	\caption{PSNR of recovered images by five methods with 50\% random element loss (iteration number is provided in parentheses)}
	\label{tab:loss_PSNR}
	\centering
	\begin{tabular}{cccccc}
		\toprule
		       Image        & TNNR  & IRNN-L$_\text{p}$ & PSSV  & Joint S$_\text{p}$ &    DW-TNNR     \\ \midrule
		\multirow{2}{*}{1}  & 22.41 &       21.60       & 24.05 &       24.00        & \textbf{25.75} \\
		                    & (283) &       (137)       & (127) &        (84)        & \textbf{(44)}  \\
		\multirow{2}{*}{2}  & 26.18 &       25.37       & 28.25 &       28.36        & \textbf{29.64} \\
		                    & (251) &       (125)       & (115) &        (92)        & \textbf{(45)}  \\
		\multirow{2}{*}{3}  & 22.64 &       21.52       & 24.01 &       24.73        & \textbf{25.37} \\
		                    & (323) &       (143)       & (136) &        (97)        & \textbf{(47)}  \\
		\multirow{2}{*}{4}  & 27.80 &       26.15       & 29.41 &       29.20        & \textbf{30.52} \\
		                    & (275) &       (130)       & (132) &        (93)        & \textbf{(46)}  \\
		\multirow{2}{*}{5}  & 26.37 &       25.84       & 28.20 &       28.52        & \textbf{29.00} \\
		                    & (279) &       (134)       & (130) &        (96)        & \textbf{(47)}  \\
		\multirow{2}{*}{6}  & 20.94 &       20.28       & 22.00 &       22.61        & \textbf{23.45} \\
		                    & (303) &       (140)       & (135) &       (103)        & \textbf{(49)}  \\
		\multirow{2}{*}{7}  & 27.98 &       26.01       & 29.62 &       29.71        & \textbf{30.48} \\
		                    & (274) &       (123)       & (115) &        (94)        & \textbf{(46)}  \\
		\multirow{2}{*}{8}  & 25.43 &       24.27       & 27.00 &       27.47        & \textbf{28.04} \\
		                    & (265) &       (126)       & (121) &       (101)        & \textbf{(46)}  \\
		\multirow{2}{*}{9}  & 21.98 &       20.40       & 23.58 &       23.43        & \textbf{24.20} \\
		                    & (278) &       (134)       & (133) &        (89)        & \textbf{(47)}  \\
		\multirow{2}{*}{10} & 23.19 &       22.47       & 25.31 &       25.76        & \textbf{26.73} \\
		                    & (294) &       (140)       & (127) &        (95)        & \textbf{(46)}  \\ \bottomrule
	\end{tabular}
\end{table}

\begin{figure}[t]
	\centering
	\subfloat[]{\includegraphics[width=0.15\textwidth]{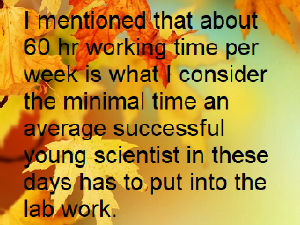}%
		\label{fig:text_incomplete}
	}
	\subfloat[]{\includegraphics[width=0.15\textwidth]{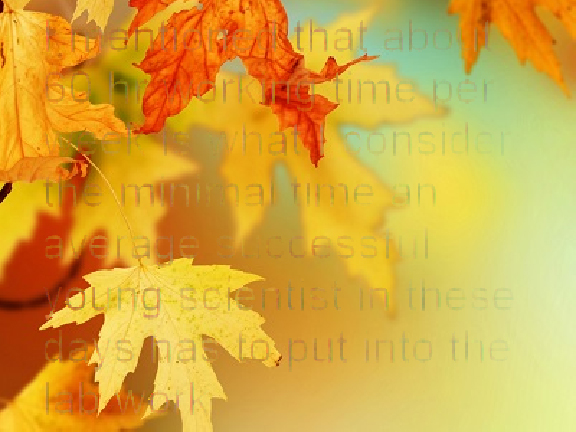}%
		\label{fig:text_TNNR}
	}
	\subfloat[]{\includegraphics[width=0.15\textwidth]{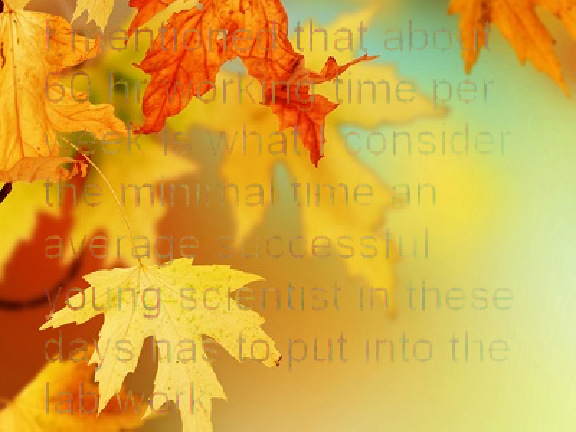}%
		\label{fig:text_IRNN}
	}
	\\[-2.0mm]
	\subfloat[]{\includegraphics[width=0.15\textwidth]{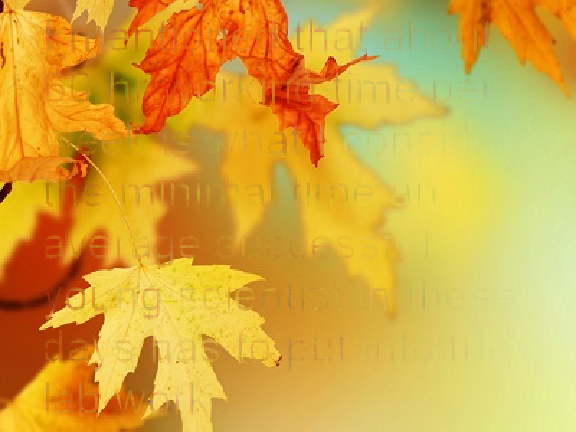}%
		\label{fig:text_PSSV}
	}
	\subfloat[]{\includegraphics[width=0.15\textwidth]{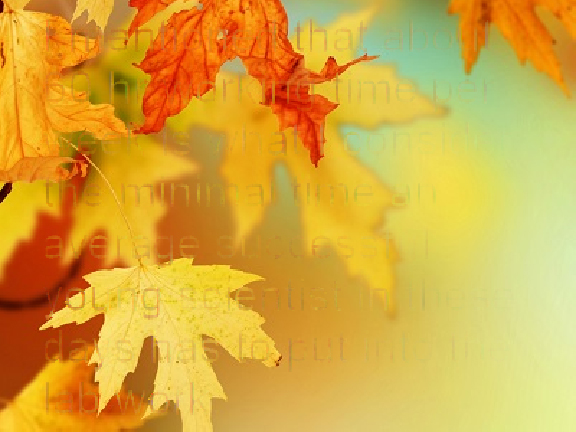}%
		\label{fig:text_Joint_Sp}
	}
	\subfloat[]{\includegraphics[width=0.15\textwidth]{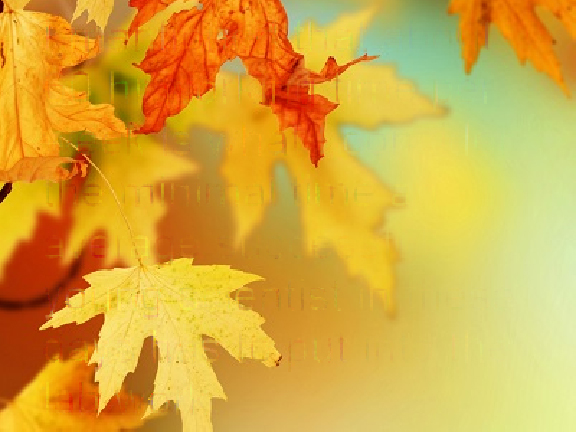}%
		\label{fig:text_DW_TNNR}
	}
	\caption{Recovered results of the 5th image in Fig.~\ref{fig:ten_images} with text block by five methods: (a) incomplete image, (b) TNNR, (c) IRNN-L$_\text{p}$, (d) PSSV, (e) Joint S$_\text{p}$, (f) DW-TNNR.}
	\label{fig:visual_text_images}
\end{figure}

2) \emph{Text block.} Text removal is quite complicated since the positions of missing elements in an image are partially contiguous. The imposed text covers a broad area rather than randomly distributed. By regarding the text as unknown elements, it is still feasible to consider it as a matrix completion problem. Let  $\theta_1 = \theta_2 = 1.2$, $\alpha_1 = 10^{-4}$, and $\rho = 1.2$ in this experiment. An example of text block is illustrated in Fig.~\ref{fig:text_incomplete}. All ten images in Fig.~\ref{fig:ten_images} are superimposed with the same text block. Then, they are restored by five approaches individually. Due to the space limit, those results of the 5th image are demonstrated in Fig.~\ref{fig:visual_text_images} for visual comparison. Table~\ref{tab:text_PSNR} provides the PSNR of recovery on all images by five methods. 

Obviously, the recovered images by the TNNR and IRNN-L$_\text{p}$ approaches (Figs.~\ref{fig:text_TNNR} and \ref{fig:text_IRNN}) consist of much more defects compared with the result of the DW-TNNR method (Fig.~\ref{fig:text_DW_TNNR}). In addition, the resulting images generated by the PSSV and Joint S$_\text{p}$ methods (Figs.~\ref{fig:text_PSSV} and \ref{fig:text_Joint_Sp}) have comparatively fewer bad pixels than the result of the TNNR. As displayed in Fig.~\ref{fig:text_DW_TNNR}, the restored image by the DW-TNNR method is visually best compared with others, and it complies with the best PSNR in Table~\ref{tab:text_PSNR}. It indicates that our DW-TNNR approach can recover images with text block.

In Table~\ref{tab:text_PSNR}, the PSNR of the DW-TNNR method obviously surpasses the corresponding results of the other methods in all cases. The PSSV and Joint S$_\text{p}$ methods run approximately to (but always inferior than) our approach. The IRNN-L$_\text{p}$ method obtains the worst results in PSNR than others. The results of the TNNR method are slightly better than IRNN-L$_\text{p}$ but clearly worse than three others. It proves that our DW-TNNR approach is quite effective to recover images with text block.

\begin{figure}[t]
	\centering
	\includegraphics[scale=0.28]{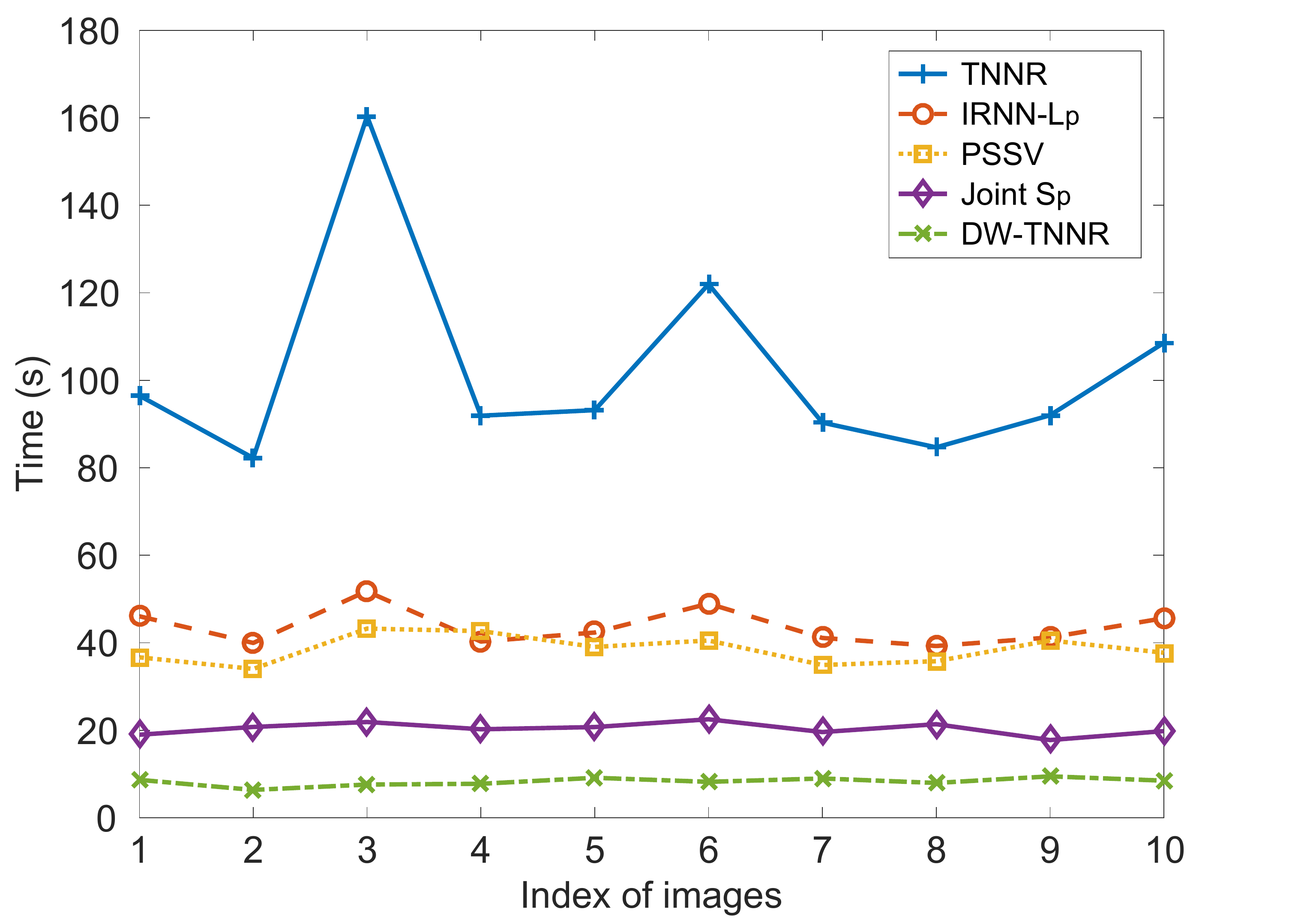}
	\caption{Time consumption by five methods on ten images in Fig.~\ref{fig:ten_images}}
	\label{fig:time_comsuming_10_images}
\end{figure}

Fig.~\ref{fig:time_comsuming_10_images} shows the elapsed time on ten images by five methods. Evidently, the DW-TNNR method runs significantly faster and more stable on each image than other methods. It consumes roughly 10 seconds to converge. The Joint S$_\text{p}$ method also has similar stability and needs nearly 20 seconds to recover images. Nevertheless, the IRNN-L$_\text{p}$ and PSSV approaches consume 35--51 seconds to accomplish the recovery tasks, whose efficiency varies on different images. Noticeably, the TNNR method runs erratically on ten images, whose elapsed time ranges from 82 to 160 seconds. Fig.~\ref{fig:time_comsuming_10_images} proves that the stability and efficiency of the DW-TNNR method are both much better than the other compared approaches. 

\begin{table}[t] \small
	\caption{PSNR of recovered images by five methods with same text block as illustrated in Fig.~\ref{fig:text_incomplete}}
	\label{tab:text_PSNR}
	\centering
	\begin{tabular}{cccccc}
		\toprule
		Image & TNNR  & IRNN-L$_\text{p}$ & PSSV  & Joint S$_\text{p}$ &    DW-TNNR     \\ \midrule
		 {1}  & 24.19 &       23.41       & 26.36 &       26.66        & \textbf{27.48} \\
		 {2}  & 26.79 &       25.87       & 28.76 &       28.11        & \textbf{29.06} \\
		 {3}  & 21.33 &       20.62       & 23.10 &       23.89        & \textbf{24.11} \\
		 {4}  & 28.82 &       27.24       & 29.98 &       30.17        & \textbf{31.03} \\
		 {5}  & 24.96 &       23.09       & 26.34 &       26.83        & \textbf{27.97} \\
		 {6}  & 21.23 &       20.20       & 23.24 &       23.05        & \textbf{24.48} \\
		 {7}  & 28.55 &       27.01       & 30.92 &       31.02        & \textbf{32.10} \\
		 {8}  & 26.04 &       25.58       & 28.32 &       28.73        & \textbf{29.22} \\
		 {9}  & 21.17 &       20.73       & 23.70 &       23.20        & \textbf{24.67} \\
		{10}  & 25.60 &       24.34       & 27.08 &       27.37        & \textbf{28.37} \\ \bottomrule
	\end{tabular}
\end{table}

\section{Conclusion} \label{sec:conclusion}

This paper proposes the double weighted truncated nuclear norm regularization (DW-TNNR) to facilitate efficient matrix completion. It applies diverse weights to the rows and columns of the matrix separately, based on the number of known elements along the corresponding dimensions, which is beneficial to the convergence with satisfactory performance. This paper develops an efficient gradient descent manner, to replace the iterative optimization of primitive TNNR method, and achieves the fast convergence with a theoretical guarantee. Furthermore, it has been confirmed that the DW-TNNR method is more robust to the number of truncated singular values than the original TNNR and other approaches. 

Extensive experimental results prove that our DW-TNNR is able to deal with two kinds of occlusions: random loss and text block. In addition, our algorithm always performs best compared with the Joint S$_\text{p}$ and PSSV approaches, and much better than the TNNR and IRNN-L$_\text{p}$ methods. In general, the DW-TNNR runs fastest in experiments under various types of loss in images. Moreover, the DW-TNNR approach is highly stable when applied to diverse scenarios, particularly compared with the TNNR and IRNN-L$_\text{p}$ approaches. In a word, our DW-TNNR has promising robustness and advantages in efficiency and accuracy for matrix completion.

\section*{Acknowledgment}


This work was supported by the Opening Foundation of the State Key Laboratory for Diagnosis and Treatment of Infectious Diseases (No.~2014KF06), the Zhejiang Provincial Natural Science Foundation of China (No.~J20130411), and the National Science and Technology Major Project (No.~2013ZX03005013).


\appendix 

\section{} \label{apdx:two_forms_of_B}
Here, we interpret two different forms of $\mathbf{B}_\ell$ in the iterative updating procedures. 

First, we recall the definition of singular value decomposition. Given $\mathbf{X} \dimension{m \times n}$, then there are two orthogonal matrices $\mathbf{U} \dimension{m \times m}$ and $\mathbf{V} \dimension{n \times n}$, which satisfies
\begin{equation}
\mathbf{X} = \mathbf{U} \Sigma \mathbf{V}^\transpose, \quad \Sigma = 
\begin{bmatrix}
\Sigma_r & \mathbf{O} \\
\mathbf{O} & \mathbf{O} \\
\end{bmatrix} ,
\end{equation}
where $\mathbf{U}=[\mathbf{u}_1,\cdots,\mathbf{u}_m]$ and $\mathbf{V}=[\mathbf{v}_1,\cdots,\mathbf{v}_n]$ are the left and right singular matrices. $\Sigma_r=\text{diag}(\sigma_1,\sigma_2,\cdots,\sigma_r)$ and $r \leq \min(m,n)$, in which those singular values $\{\sigma_i\}_{i=1}^{r} > 0$ are in decreasing order. 

According to the Von Neumann's trace inequality, we obtain
\begin{gather}
\trace{\mathbf{A}\mathbf{X}\mathbf{B}^\transpose}  \leq \norm{\mathbf{X}}_* = \sum_{i=1}^{r} \sigma_i = \trace{\Sigma_r} , \\
\begin{aligned} \label{eq:AXB=UrXVr}
\max_{\mathbf{A}\mathbf{A}^\transpose=\mathbf{B}\mathbf{B}^\transpose=\mathbf{I}}  \trace{\mathbf{A}\mathbf{X}\mathbf{B}^\transpose} &= \trace{\Sigma_r} \\
&= \trace{[\mathbf{u}_1,\cdots,\mathbf{u}_r]^\transpose \mathbf{X} [\mathbf{v}_1,\cdots,\mathbf{v}_r]}	\\
&= \trace{\mathbf{U}_r^\transpose \mathbf{X} \mathbf{V}_r} ,
\end{aligned}
\end{gather}
where $\mathbf{A} \dimension{m \times m}$ and $\mathbf{B} \dimension{m \times n}$, the sizes of which are much larger than $\mathbf{U}_r \dimension{m \times r}$ and $\mathbf{V}_r \dimension{n \times r}$, respectively. Hence, the dimensions of $\mathbf{A}\mathbf{X}\mathbf{B}^\transpose$ and $\mathbf{U}_r^\transpose \mathbf{X} \mathbf{V}_r$ in \eqref{eq:AXB=UrXVr} are not equal if without the trace function. Notice that $\Sigma = \mathbf{U}^\transpose \mathbf{X} \mathbf{V} \dimension{m \times n}$ is not square. Unfortunately, the trace operator only makes sense when acting on a square matrix. 

\begin{figure}[t]
	\centering
	\includegraphics[width=3in]{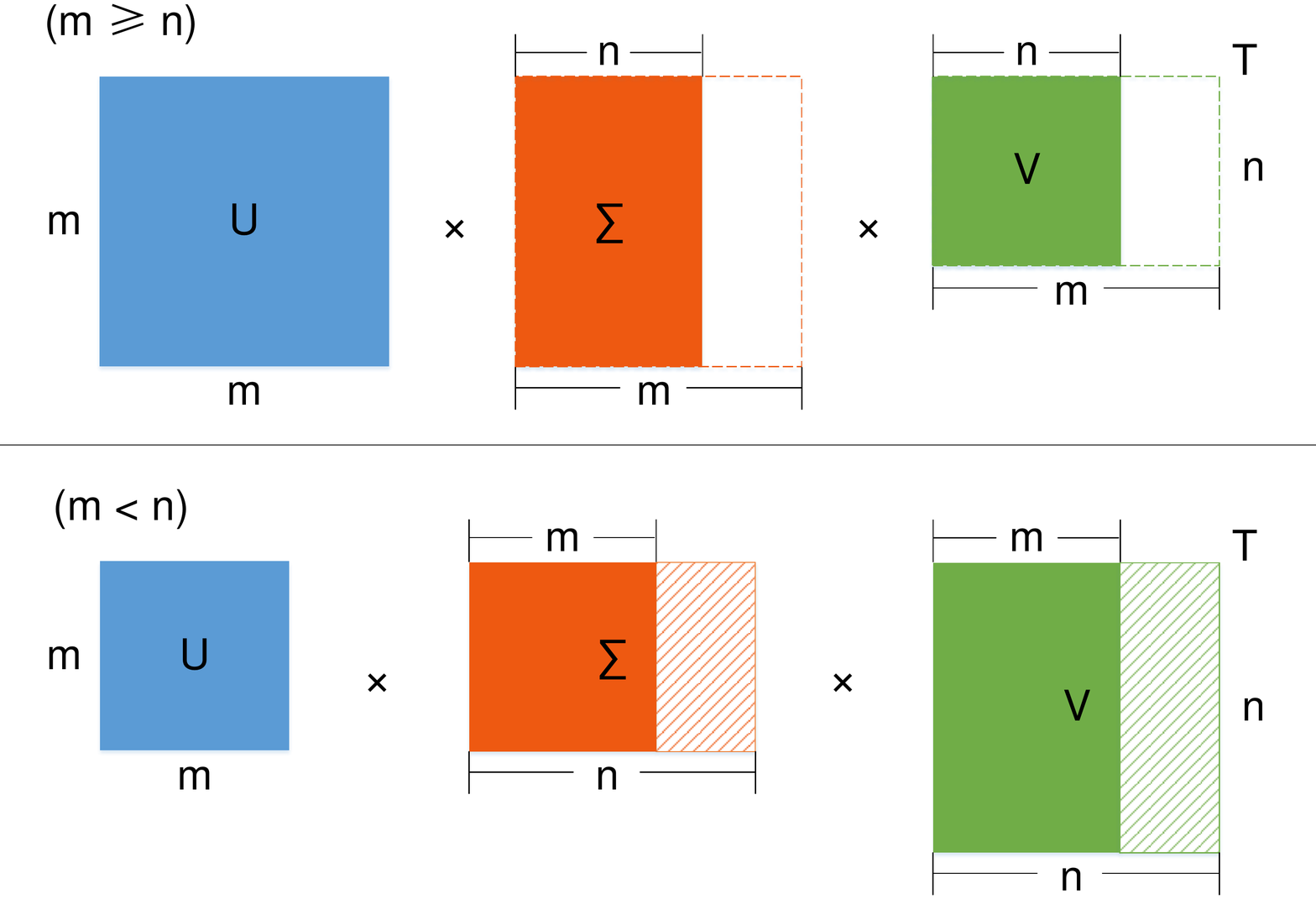}
	\caption{Two cases of $\mathbf{X}$'s singular value decomposition. $\mathbf{V}$ requires to be extended ($m \geq n$) or trimmed ($m < n$), for the consistence with the dimension of $\mathbf{B}^\transpose$.}
	\label{fig:appendix_A}
\end{figure}

The dimension of $\mathbf{U}^\transpose$ is consistent with the counterpart $\mathbf{A}$ while the dimension of $\mathbf{V}$ is inconsistent with the counterpart of $\mathbf{B}^\transpose$. Thus, $\mathbf{V}$ requires to be altered in the light of two cases, $m \geq n$ or $m < n$. Fig.~\ref{fig:appendix_A} clarifies the process intuitively. If $m \geq n$, $\mathbf{V}$ should be extended with zeros along the column dimension. Otherwise, it should be trimmed along the column dimension. Mathematically, we conclude
\begin{equation}
\mathbf{B} = 
\begin{cases}
[\mathbf{V}, \mathbf{O}_{n \times (m-n)}]^\transpose , & m \geq n , \\
[\mathbf{v}_1,\cdots,\mathbf{v}_m]^\transpose, & m < n. \\
\end{cases}
\end{equation}

\section{} \label{apdx:proof_X_update}

Beforehand, let us recall updating procedures \eqref{eq:update_W}--\eqref{eq:update_Y} in Step 2, written by
\begin{equation}
\mathbf{W}_{t+1} = \mathbf{X}_t + \frac{1}{\mu_t} ( \mathbf{P}^{-2} \mathbf{C}_k^\transpose \mathbf{D}_k \mathbf{Q}^{-2} + \mathbf{P}^{-1} \mathbf{Y}_t \mathbf{Q}^{-1} ) , \label{eq:apdx_update_W1}	
\end{equation}
\begin{equation}
\mathbf{W}_{t+1} = (\mathbf{W}_{t+1})_{\Omega^{\text{c}}} + \mathbf{M}_{\Omega}, \label{eq:apdx_update_W2}
\end{equation}
\begin{equation}
\mathbf{X}_{t+1} = \mathbf{W}_{t+1} - \frac{1}{\mu_t} ( \mathbf{P}^{-2} \mathbf{A}_k^\transpose \mathbf{B}_k \mathbf{Q}^{-2} + \mathbf{P}^{-1} \mathbf{Y}_t \mathbf{Q}^{-1} ) , \label{eq:apdx_update_X}
\end{equation}
\begin{equation}
\mathbf{Y}_{t+1} = \mathbf{Y}_t + \beta_t \mathbf{P}(\mathbf{X}_{t+1}-\mathbf{W}_{t+1})\mathbf{Q} , \label{eq:apdx_update_Y}
\end{equation}
where $\mathbf{A}_k$, $\mathbf{B}_k$, $\mathbf{C}_k$, and $\mathbf{D}_k$ are obtained in the $k$-th iteration of Step 1, by $\mathbf{X}_k$'s singular value decomposition. $\mathbf{P}$ and $\mathbf{Q}$ are weight matrices applied for rows and columns.  $\mu_t$ and $\beta_t$ are both positive scalars.

In advance, we clarify two norm inequalities:
\begin{align}
\norm{\mathbf{X} + \mathbf{Y}}_\fro & \leq \norm{\mathbf{X}}_\fro + \norm{\mathbf{Y}}_\fro,\\
\norm{\mathbf{X} \cdot \mathbf{Y}}_\fro & \leq \norm{\mathbf{X}}_\fro \cdot \norm{\mathbf{Y}}_\fro.
\end{align}
Note $\forall \, k = 1, 2, \cdots$, $\norm{\mathbf{A}_k^\transpose \mathbf{B}_k}_\fro = \sqrt{m}$ and $\norm{\mathbf{C}_k^\transpose \mathbf{D}_k}_\fro = \sqrt{r}$.

By substituting \eqref{eq:apdx_update_W1} into \eqref{eq:apdx_update_X}, we obtain
\begin{align}
\mathbf{X}_{t+1} &= \mathbf{X}_t  - \frac{1}{\mu_t} \mathbf{P}^{-2} (\mathbf{A}_k^\transpose \mathbf{B}_k - \mathbf{C}_k^\transpose \mathbf{D}_k) \mathbf{Q}^{-2} , \\
\norm{\mathbf{X}_{t+1} - \mathbf{X}_t}_\fro &= \frac{1}{\mu_t} \norm{ \mathbf{P}^{-2} (\mathbf{A}_k^\transpose \mathbf{B}_k - \mathbf{C}_k^\transpose \mathbf{D}_k) \mathbf{Q}^{-2} }_\fro \nonumber \\
& \leq \frac{1}{\mu_t} \norm{ \mathbf{P}^{-2} }_\fro \left ( \norm{ \mathbf{A}_k^\transpose \mathbf{B}_k }_\fro + \norm{ \mathbf{C}_k^\transpose \mathbf{D}_k }_\fro \right ) \norm{ \mathbf{Q}^{-2} }_\fro \nonumber \\
& \leq \frac{1}{\mu_t} \norm{ \mathbf{P}^{-2} }_\fro \, ( \sqrt{m} + \sqrt{r}  ) \, \norm{ \mathbf{Q}^{-2} }_\fro. 
\end{align}
If $\lim\limits_{t \rightarrow \infty} \frac{1}{\mu_t} = 0$, i.e. $\mu_t$ increases progressively, we have
\begin{equation}
\lim_{t \rightarrow \infty} \norm{\mathbf{X}_{t+1} - \mathbf{X}_t}_\fro = 0.
\end{equation}
Therefore, after numerous iterations, $\mathbf{X}_t$ will converge. Due to the convexity of unconstrained Lagrangian function \eqref{eq:Weighted_TNNR_Lagrangian} with respect to $\mathbf{X}$ and $\mathbf{W}$, $\mathbf{X}_t$ is confirmed to reach its local optimal solution eventually.

Suppose it consumes $N$ iterations to converge, we further conclude that
\begin{align}
\mathbf{X}_N 
	&= \mathbf{X}_{N-1}  - \frac{1}{\mu_{N-1}} \mathbf{P}^{-2} (\mathbf{A}_k^\transpose \mathbf{B}_k - \mathbf{C}_k^\transpose \mathbf{D}_k) \mathbf{Q}^{-2} \nonumber \\
	&= \mathbf{X}_{N-2}  - \left (\frac{1}{\mu_{N-1}} + \frac{1}{\mu_{N-2}} \right ) \mathbf{P}^{-2} (\mathbf{A}_k^\transpose \mathbf{B}_k - \mathbf{C}_k^\transpose \mathbf{D}_k) \mathbf{Q}^{-2} \nonumber \\
	& \ \ \vdots \nonumber \\
	&= \mathbf{X}_1 - \sum_{t=1}^{N-1} \frac{1}{\mu_t} \mathbf{P}^{-2} (\mathbf{A}_k^\transpose \mathbf{B}_k - \mathbf{C}_k^\transpose \mathbf{D}_k) \mathbf{Q}^{-2} ,	
\end{align}
where $\mathbf{X}_1 = \mathbf{X}_k$ is the result of Step 1 in the $k$-th iteration, and $\mathbf{X}_1$ is regarded as the initial value of the iterative optimization scheme in Step 2.

By incorporating \eqref{eq:apdx_update_W2} and the constraint $\mathbf{X} = \mathbf{W}$, we have
\begin{gather}
\mathbf{X}_N = \mathbf{X}_1 - \frac{1}{\alpha_k} \mathbf{P}^{-2} (\mathbf{A}_k^\transpose \mathbf{B}_k - \mathbf{C}_k^\transpose \mathbf{D}_k) \mathbf{Q}^{-2} , \\
\mathbf{X}_{k+1} = (\mathbf{X}_N)_{\Omega^{\text{c}}} + \mathbf{M}_{\Omega},
\end{gather}
where $\mathbf{X}_1 = \mathbf{X}_k$ and $\frac{1}{\alpha_k} = \sum_{t=1}^{N-1} \frac{1}{\mu_t}$. Therefore, the proof of Theorem~\ref{thm:X_update} is finished.

\section{} \label{apdx:proof_alpha}

First, we rewrite \eqref{eq:update_X_simple_gradient} in Section~\ref{sec:proposed_method} as follows: 
\begin{equation}
\mathbf{X}_{k+1} - \mathbf{X}_k = - \frac{1}{\alpha_k} \mathcal{P} \mathbf{\varPhi}_k^\transpose \mathbf{\varLambda}_k \mathcal{Q}.
\end{equation}
Next, we evaluate it through
\begin{align}
\norm{ \mathbf{X}_{k+1} - \mathbf{X}_k }_\fro &= \frac{1}{\alpha_k} \norm{ \mathcal{P} \mathbf{\varPhi}_k^\transpose \mathbf{\varLambda}_k \mathcal{Q} }_\fro \nonumber \\
&\leq \frac{1}{\alpha_k} \norm{\mathcal{P}}_\fro \, \norm{ \mathbf{\varPhi}_k^\transpose \mathbf{\varLambda}_k }_\fro \, \norm{\mathcal{Q}}_\fro \nonumber \\
&\leq \frac{1}{\alpha_k} \norm{\mathcal{P}}_\fro \, \norm{ \mathbf{A}_k^\transpose \mathbf{B}_k - \mathbf{C}_k^\transpose \mathbf{D}_k }_\fro \, \norm{\mathcal{Q}}_\fro \nonumber \\
&\leq \frac{1}{\alpha_k} \norm{\mathcal{P}}_\fro \, (\sqrt{m} + \sqrt{r}) \, \norm{\mathcal{Q}}_\fro, 
\end{align}
where two norm inequalities are used, $m$ denotes the number of rows of $\mathbf{X}$, and $r$ denotes the number of truncated singular values.

Let $\varepsilon$ indicate a tolerance as the stopping criterion of the DW-TNNR method. We have
\begin{equation}
\frac{1}{\alpha_k} \norm{\mathcal{P}}_\fro \, (\sqrt{m} + \sqrt{r}) \, \norm{\mathcal{Q}}_\fro \leq \varepsilon. \label{eq:apdx_epsilon}
\end{equation}
In addition, we define that $\alpha$ is scaled by $\rho$, i.e.~$\alpha_{k+1} = \rho \, \alpha_k$. So we have the following equation: 
\begin{equation}
\alpha_{k} = \rho \, \alpha_{k-1} = \rho^2  \alpha_{k-2} = \cdots = \rho^{k-1}  \alpha_1.
\end{equation}
To simplify the notation, let $\gamma = \norm{\mathcal{P}}_\fro \, (\sqrt{m} + \sqrt{r}) \, \norm{\mathcal{Q}}_\fro$. Thus, \eqref{eq:apdx_epsilon} can be reformulated as 
\begin{gather}
\frac{1}{\rho^{k-1} \, \alpha_1} \gamma \leq \varepsilon , \\
k \geq 1 + \frac{\ln \gamma - \ln (\alpha_1 \varepsilon)}{\ln \rho} .
\end{gather}
Thus, the proof of Theorem~\ref{thm:alpha_proof} is accomplished.




\begin{thebibliography}{10}
	\expandafter\ifx\csname url\endcsname\relax
	\def\url#1{\texttt{#1}}\fi
	\expandafter\ifx\csname urlprefix\endcsname\relax\def\urlprefix{URL }\fi
	\expandafter\ifx\csname href\endcsname\relax
	\def\href#1#2{#2} \def\path#1{#1}\fi
	
	\bibitem{Hu2017-Motion}
	W.~Hu, Z.~Wang, S.~Liu, X.~Yang, G.~Yu, J.~Zhang, Motion capture data
	completion via truncated nuclear norm regularization, IEEE Signal Processing
	Letters PP~(99) (2017) 1--1.
	\newblock \href {http://dx.doi.org/10.1109/LSP.2017.2687044}
	{\path{doi:10.1109/LSP.2017.2687044}}.
	
	\bibitem{Adeli2015-Non-negative}
	E.~Adeli-Mosabbeb, M.~Fathy, Non-negative matrix completion for action
	detection, Image and Vision Computing 39~(Supplement C) (2015) 38--51.
	\newblock \href {http://dx.doi.org/10.1016/j.imavis.2015.04.006}
	{\path{doi:10.1016/j.imavis.2015.04.006}}.
	
	\bibitem{Li2015-Non-Local}
	W.~Li, L.~Zhao, Z.~Lin, D.~Xu, D.~Lu, Non-local image inpainting using low-rank
	matrix completion, Computer Graphics Forum 34~(6) (2015) 111--122.
	\newblock \href {http://dx.doi.org/10.1111/cgf.12521}
	{\path{doi:10.1111/cgf.12521}}.
	
	\bibitem{Hu2013-Accurate}
	Y.~Hu, D.~Zhang, J.~Ye, X.~Li, X.~He, Fast and accurate matrix completion via
	truncated nuclear norm regularization, IEEE Transactions on Pattern Analysis
	and Machine Intelligence 35~(9) (2013) 2117--2130.
	\newblock \href {http://dx.doi.org/10.1109/TPAMI.2012.271}
	{\path{doi:10.1109/TPAMI.2012.271}}.
	
	\bibitem{Luo2015-Multiview}
	Y.~Luo, T.~Liu, D.~Tao, C.~Xu, Multiview matrix completion for multilabel image
	classification, IEEE Transactions on Image Processing 24~(8) (2015)
	2355--2368.
	\newblock \href {http://dx.doi.org/10.1109/TIP.2015.2421309}
	{\path{doi:10.1109/TIP.2015.2421309}}.
	
	\bibitem{Mansour2014-Video}
	H.~Mansour, A.~Vetro, Video background subtraction using semi-supervised robust
	matrix completion, in: IEEE International Conference on Acoustics, Speech and
	Signal Processing, 2014, pp. 6528--6532.
	\newblock \href {http://dx.doi.org/10.1109/icassp.2014.6854862}
	{\path{doi:10.1109/icassp.2014.6854862}}.
	
	\bibitem{Yang2014-Background}
	J.~Yang, X.~Sun, X.~Ye, K.~Li, Background extraction from video sequences via
	motion-assisted matrix completion, in: IEEE International Conference on Image
	Processing, 2014, pp. 2437--2441.
	\newblock \href {http://dx.doi.org/10.1109/ICIP.2014.7025493}
	{\path{doi:10.1109/ICIP.2014.7025493}}.
	
	\bibitem{Lee2016-Computationally}
	C.~Lee, E.~Lam, Computationally efficient truncated nuclear norm minimization
	for high dynamic range imaging, IEEE Transactions on Image Processing 25~(9)
	(2016) 4145--4157.
	\newblock \href {http://dx.doi.org/10.1109/TIP.2016.2585047}
	{\path{doi:10.1109/TIP.2016.2585047}}.
	
	\bibitem{Candes2010-Matrix}
	E.~J. Cand\`{e}s, Y.~Plan, Matrix completion with noise, Proceedings of the
	IEEE 98~(6) (2010) 925--936.
	\newblock \href {http://dx.doi.org/10.1109/JPROC.2009.2035722}
	{\path{doi:10.1109/JPROC.2009.2035722}}.
	
	\bibitem{Candes2010-Power}
	E.~J. Cand\`{e}s, T.~Tao, The power of convex relaxation: Near-optimal matrix
	completion, IEEE Transactions on Information Theory 56~(5) (2010) 2053--2080.
	\newblock \href {http://dx.doi.org/10.1109/TIT.2010.2044061}
	{\path{doi:10.1109/TIT.2010.2044061}}.
	
	\bibitem{Candes2009-Exact}
	E.~J. Cand\`{e}s, B.~Recht, Exact matrix completion via convex optimization,
	Foundations of Computational Mathematics 9~(6) (2009) 717--772.
	\newblock \href {http://dx.doi.org/10.1007/s10208-009-9045-5}
	{\path{doi:10.1007/s10208-009-9045-5}}.
	
	\bibitem{Liu2016-Low-Rank}
	G.~Liu, P.~Li, Low-rank matrix completion in the presence of high coherence,
	IEEE Transactions on Signal Processing 64~(21) (2016) 5623--5633.
	\newblock \href {http://dx.doi.org/10.1109/TSP.2016.2586753}
	{\path{doi:10.1109/TSP.2016.2586753}}.
	
	\bibitem{Candes2011-Robust}
	E.~J. Cand\`{e}s, X.~Li, Y.~Ma, J.~Wright, Robust principal component
	analysis?, Journal of the ACM 58~(3) (2011) 11:1--11:37.
	\newblock \href {http://dx.doi.org/10.1145/1970392.1970395}
	{\path{doi:10.1145/1970392.1970395}}.
	
	\bibitem{Wright2009-Robust}
	J.~Wright, A.~Ganesh, S.~Rao, Y.~Peng, Y.~Ma, Robust principal component
	analysis: Exact recovery of corrupted low-rank matrices via convex
	optimization, in: Advances in Neural Information Processing Systems, 2009,
	pp. 2080--2088.
	
	\bibitem{Cai2010-Singular}
	J.~Cai, E.~J. Cand\`{e}s, Z.~Shen, A singular value thresholding algorithm for
	matrix completion, SIAM Journal on Optimization 20~(4) (2010) 1956--1982.
	\newblock \href {http://dx.doi.org/10.1137/080738970}
	{\path{doi:10.1137/080738970}}.
	
	\bibitem{Lin2011-Linearized}
	Z.~Lin, R.~Liu, Z.~Su, Linearized alternating direction method with adaptive
	penalty for low-rank representation, in: Advances in Neural Information
	Processing Systems, 2011, pp. 612--620.
	
	\bibitem{Toh2010-accelerated}
	K.~C. Toh, S.~Yun, An accelerated proximal gradient algorithm for nuclear norm
	regularized linear least squares problems, Pacific Journal of Optimization
	6~(15) (2010) 615--640.
	
	\bibitem{Lu2016-Nonconvex}
	C.~Lu, J.~Tang, S.~Yan, Z.~Lin, Nonconvex nonsmooth low rank minimization via
	iteratively reweighted nuclear norm, IEEE Transactions on Image Processing
	25~(2) (2016) 829--839.
	\newblock \href {http://dx.doi.org/10.1109/TIP.2015.2511584}
	{\path{doi:10.1109/TIP.2015.2511584}}.
	
	\bibitem{Li2017-Weighted}
	C.~Li, X.~Wang, L.~Zhang, J.~Tang, H.~Wu, L.~Lin, Weighted low-rank
	decomposition for robust grayscale-thermal foreground detection, IEEE
	Transactions on Circuits and Systems for Video Technology 27~(4) (2017)
	725--738.
	\newblock \href {http://dx.doi.org/10.1109/TCSVT.2016.2556586}
	{\path{doi:10.1109/TCSVT.2016.2556586}}.
	
	\bibitem{Song2016-Image}
	W.~Song, J.~Zhu, Y.~Li, C.~Chen, Image alignment by online robust {PCA} via
	stochastic gradient descent, IEEE Transactions on Circuits and Systems for
	Video Technology 26~(7) (2016) 1241--1250.
	\newblock \href {http://dx.doi.org/10.1109/TCSVT.2015.2455711}
	{\path{doi:10.1109/TCSVT.2015.2455711}}.
	
	\bibitem{Xue2017-Robust}
	S.~Xue, X.~Jin, Robust classwise and projective low-rank representation for
	image classification, Signal, Image and Video Processing 1~(1) (2017) 1--9.
	\newblock \href {http://dx.doi.org/10.1007/s11760-017-1136-1}
	{\path{doi:10.1007/s11760-017-1136-1}}.
	
	\bibitem{Bhojanapalli2014-Universal}
	S.~Bhojanapalli, P.~Jain, Universal matrix completion, in: 31st International
	Conference on Machine Learning, Bejing, China, 2014, pp. 1881--1889.
	
	\bibitem{Chen2014-Coherent}
	Y.~Chen, S.~Bhojanapalli, S.~Sanghavi, R.~Ward, Coherent matrix completion, in:
	31st International Conference on Machine Learning, Bejing, China, 2014, pp.
	674--682.
	
	\bibitem{Oh2016-Partial}
	T.~H. Oh, Y.~W. Tai, J.~C. Bazin, H.~Kim, I.~S. Kweon, Partial sum minimization
	of singular values in robust {PCA}: Algorithm and applications, IEEE
	Transactions on Pattern Analysis and Machine Intelligence 38~(4) (2016)
	744--758.
	\newblock \href {http://dx.doi.org/10.1109/TPAMI.2015.2465956}
	{\path{doi:10.1109/TPAMI.2015.2465956}}.
	
	\bibitem{Nie2015-Joint}
	F.~Nie, H.~Wang, H.~Huang, C.~Ding, Joint {Schatten} $p$-norm and $\ell_p$-norm
	robust matrix completion for missing value recovery, Knowledge and
	Information Systems 42~(3) (2015) 525--544.
	\newblock \href {http://dx.doi.org/10.1007/s10115-013-0713-z}
	{\path{doi:10.1007/s10115-013-0713-z}}.
	
	\bibitem{Hong2016-Online}
	B.~Hong, L.~Wei, Y.~Hu, D.~Cai, X.~He, Online robust principal component
	analysis via truncated nuclear norm regularization, Neurocomputing 175, Part
	A (2016) 216--222.
	\newblock \href {http://dx.doi.org/10.1016/j.neucom.2015.10.052}
	{\path{doi:10.1016/j.neucom.2015.10.052}}.
	
	\bibitem{Feng2013-Online}
	J.~Feng, H.~Xu, S.~Yan, Online robust {PCA} via stochastic optimization, in:
	Advances in Neural Information Processing Systems, 2013, pp. 404--412.
	
	\bibitem{Hu2015-Large}
	Y.~Hu, Z.~Jin, Y.~Shi, D.~Zhang, D.~Cai, X.~He, Large scale multi-class
	classification with truncated nuclear norm regularization, Neurocomputing 148
	(2015) 310--317.
	\newblock \href {http://dx.doi.org/10.1016/j.neucom.2014.06.073}
	{\path{doi:10.1016/j.neucom.2014.06.073}}.
	
	\bibitem{Cao2017-Recovering}
	F.~Cao, J.~Chen, H.~Ye, J.~Zhao, Z.~Zhou, Recovering low-rank and sparse matrix
	based on the truncated nuclear norm, Neural Networks 85 (2017) 10--20.
	\newblock \href {http://dx.doi.org/10.1016/j.neunet.2016.09.005}
	{\path{doi:10.1016/j.neunet.2016.09.005}}.
	
	\bibitem{Neumann1937-matrix}
	J.~Neumann, Some matrix inequalities and metrization of matrix space, Tomsk
	Univ. Rev 1~(11) (1937) 286--300.
	
\end{thebibliography}


\end{document}